\documentclass[sigconf]{acmart}

\RequirePackage{OpenProc-Mods}

\AddToHook{cmd/maketitle/before}{%

\def\EDBTISSN{}
\def\EDBTISBN{}
\setcopyright{none}
\copyrightyear{}
\acmYear{2026}
\acmDOI{}
\acmISBN{}

\acmConference{}

\newsavebox{\ximagebox}
\newlength{\ximageheight}
\newsavebox{\xglyphbox}
\newlength{\xglyphheight}
\newcommand{\xbox}[1]%
  {\savebox{\ximagebox}{#1}%
  \settoheight{\ximageheight}{\usebox{\ximagebox}}%
  \savebox{\xglyphbox}{\color{white}\char32}%
  \settoheight{\xglyphheight}{\usebox{\xglyphbox}}%
  \raisebox{\ximageheight}[0pt][0pt]{\raisebox{-\xglyphheight}[0pt][0pt]{%
    \makebox[0pt][l]{\usebox{\xglyphbox}}}}%
    \usebox{\ximagebox}%
    \raisebox{0pt}[0pt][0pt]{\makebox[0pt][r]{\usebox{\xglyphbox}}}}

\newsavebox{\LogoBox}

\fancypagestyle{firstpagestyle}{%
  \fancyhf{}%
  \fancyfoot{}%
  \fancyhead[L,C]{}
  \fancyhead[R]{\href{https://OpenProceedings.org/}{\raisebox{15pt}[0pt][0pt]{\xbox{\usebox{\LogoBox}}}}}
  }
  
\hypersetup{%
  pdfcopyright={CC-by-nc-nd},
  pdfpublisher={OpenProceedings.org -- University of Konstanz, Germany},
  pdfcreator={LaTeX with acmart, hyperref, and EDBT modifications},
  pdfvolumenum={30},
  pdfissuenum={1},
  pdfisbn={\EDBTISBN},
  pdfissn={\EDBTISSN},
  pdfdoi={10.48786/edbt.2027.xx}
}
}

\def\BibTeX{{\rm B\kern-.05em{\sc i\kern-.025em b}\kern-.08em
    T\kern-.1667em\lower.7ex\hbox{E}\kern-.125emX}}

\usepackage{multirow}
\usepackage{makecell}

\begin{document}


\OPtitle{SINT-Flow: Schema Integration using Large Language Model Workflows}



\OPauthor[0000-0002-2158-0070]{Keti Korini}
\email{kkorini@uni-mannheim.de}
\affiliation{%
  \institution{University of Mannheim}
  \city{Mannheim}
  \country{Germany}
}

\OPauthor[0000-0003-2367-0237]{Christian Bizer}
\email{christian.bizer@uni-mannheim.de}
\affiliation{%
  \institution{University of Mannheim}
  \city{Mannheim}
  \country{Germany}}


\renewcommand{\shortauthors}{Korini and Bizer} 


\begin{OPabstract}
  The goal of schema integration is, given a set of input schemata or tables, to derive a global, unified schema that is able to represent the concepts, attributes, and relationships of all input tables in a coherent fashion. This paper presents SINT-Flow, a schema integration framework composed of five LLM-based operators that can be combined into workflows to perform fully automated, end-to-end schema integration. In contrast to existing approaches, SINT-Flow can process denormalized source tables that contain attributes describing multiple entity types. During the schema integration process, these tables are decomposed into separate entity-specific relations. To evaluate SINT-Flow, we introduce SINT-Bench, a schema integration benchmark comprising 10 schema integration tasks consisting of altogether 93 relational tables, including tables that describe multiple types of entities. We evaluate SINT-Flow using GPT-5.2 as well as the open-weight model Qwen-3.6-27B as alternative backbone models. Using these models, SINT-Flow achieves a Completeness rate of at least 97\% in entity type detection, 81\% for attribute detection, and an F1-score of 84\% for schema mapping. Furthermore, we perform an ablation study to prove the utility of the applied self-consistency strategy as well as the inclusion of a review loop into the Schema Matching Operator.
\end{OPabstract}


\maketitle

\section{Introduction}
\label{sec:introduction}

Many applications in decision-making, data analysis, and machine learning require the integration of data from multiple heterogeneous sources as part of the data preparation process.
Two basic data integration scenarios can be distinguished with respect to the target schema to which the source tables are mapped. 
In the first scenario, the target schema is determined by the use case at hand, e.g., the specific goal of the use case determines which types of entities and which attributes describing these entities are relevant and therefore should be part of the target schema. The heterogeneous input schemata are mapped to this use-case-specific target schema. In the second scenario, data from heterogeneous sources are integrated into a single representation, which is subsequently used for multiple, not yet fully known use cases. In order not to lose potentially useful data, in this scenario, a \textit{global, unified schema} that can represent all data from the sources is derived from the input tables. This task is called \textit{schema integration}~\cite{batini1986comparative,bernstein2002schemamanagement}. According to Batini et al.~\cite{batini1986comparative}, the integrated schema should fulfill the following requirements: \textbf{1) Completeness and Correctness}: The integrated schema should correctly represent all concepts contained in any of the source schemata.
\textbf{2) Minimality}: A concept occurring in multiple source schemata should be represented only once in the integrated schema.
\textbf{3) Understandability}: The integrated schema should be easy to understand in order to facilitate the use of the integrated data for downstream tasks. In this paper, we investigate the variant of the schema integration task in which the input consists solely of a set of tables containing column headers and attribute values, without any additional metadata or documentation.

\begin{figure}
  \centering
  \includegraphics[width=\columnwidth]{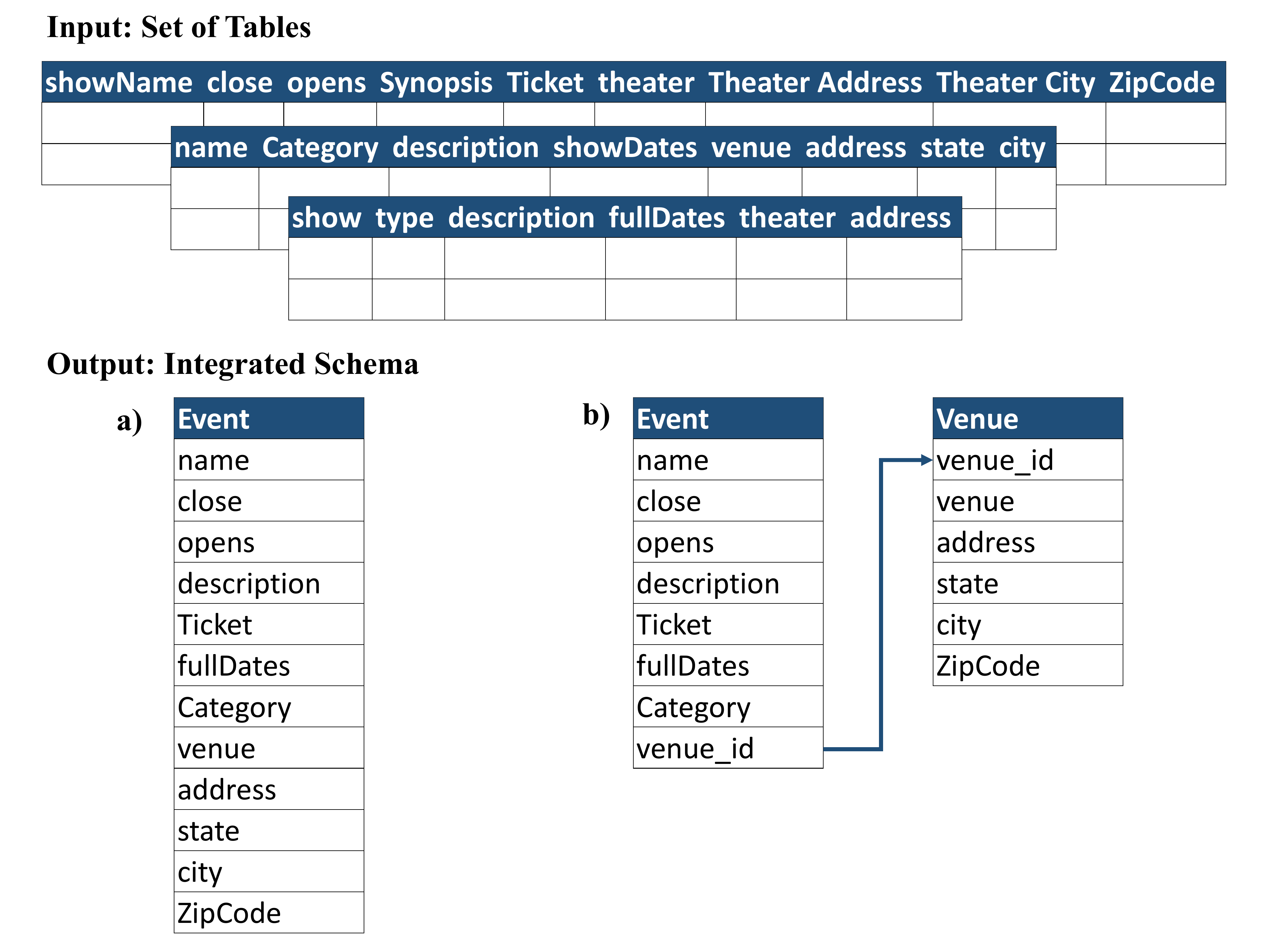}
  \caption{Example of a schema integration task consisting of a set of input tables together with two alternative integrated schemata that can be derived from the input tables: a) flat global schema; b) entity-specific global schema.}
  \label{fig:schema-integration-task}
\end{figure}

Figure~\ref{fig:schema-integration-task} illustrates a schema integration task in which the input consists of a collection of relational tables describing events and their associated venues. These tables are shown in the upper part of the figure. The lower part presents two alternative integrated schemata that can be derived from the source tables.
The first alternative, shown in Figure~\ref{fig:schema-integration-task}a, is obtained by identifying attribute correspondences across the source schemata and grouping semantically equivalent attributes. The resulting flat global schema contains the union of the source attributes while representing each distinct concept only once.
The source tables, however, combine attributes describing two different entity types, namely \textit{Event} and \textit{Venue}. Thus, the sources can alternatively be integrated using the integrated schema shown in Figure~\ref{fig:schema-integration-task}b. This schema decomposes the data into two separate types, \textit{Event} and \textit{Venue}, linked through a generated venue identifier used as a foreign key. Such a decomposition reduces data redundancy.

There exists a large body of related research on schema integration~\cite{batini1986comparative,bernstein2002schemamanagement}. Earlier methods from the 1980s and 1990s are often semi-automatic and combine integration rules with human involvement~\cite{spaccapietra1992,navathe1982methodology,batini1984methodology,batini1986comparative}. In the 2000s, fully automatic probabilistic schema integration methods were introduced which exploit attribute co-occurence~\cite{das2008bootstrapping, he2003statistical,he2006automatic,su2006holistic,chuang2008integrating}. Other works use clustering algorithms to derive the integrated schemata~\cite{pei2006novel,khatiwada2022integrating,wu2025schema}. In 2025, initial research started to use large language models (LLMs) to discover hierarchical integrated schemata ~\cite{wu2025taxonomy,wu2025schema}, but without splitting input tables that describe multiple entity types.

\textbf{Research gap:} To the best of our knowledge, existing schema integration methods are either semi-automatic or do not split input tables that describe multiple entity types as they do not distinguish which attributes within the same table refer to which entity type.
In this paper, we investigate the use of LLM-based workflows to derive entity-specific integrated schemata from input tables by assigning attributes within each input table to an entity type.  
We construct LLM workflows consisting of five LLM-based operators: the \textit{Table Splitting}, \textit{Schema Matching}, \textit{Table Grouping}, \textit{Attribute Merging}, and \textit{Integrated Schema Output} operators. We test the workflows using two LLMs as backbone models: one hosted LLM (OpenAI's GPT-5.2) and an open-weight model (Qwen-3.6-27B) to cover the use case where local data processing is required due to legal constraints. To compare with the related work, we test part of our workflow on a collection of real-world tables from the Real Benchmark~\cite{khatiwada2022integrating}. We evaluate the full workflows on SINT-Bench, a schema integration benchmark consisting of 10 tasks that we create and that includes integrated schemata for multiple entity types. We further show how we apply a self-consistency strategy~\cite{wang2023selfconsistency} to build consistent operator outputs from multiple runs. Our method derives integrated schemata from a set of source tables. It does not translate the data from the source tables into the derived integrated schema and does not include record deduplication.

\textbf{Contributions:} 
The contributions of the paper are:
\begin{enumerate}
    \item We introduce SINT-Flow, a schema integration framework consisting of five LLM-based operators which can be combined into different sequences to form schema integration workflows. In the paper, we propose three workflows. What distinguishes these workflows from existing approaches is their ability to split tables describing multiple types of entities into separate entity tables and derive the integrated schemata for the detected entity types.
    
    \item We introduce SINT-Bench, a schema integration benchmark consisting of 10 schema integration tasks with altogether 93 tables. Eight of the tasks involve two or more entity types and require the splitting of the columns of each input table per entity type. Compared to existing benchmarks, SINT-Bench is the first schema integration benchmark to include ground truth for the detection of entity types and their attributes within each input table.

    \item We evaluate the proposed workflows using SINT-Bench. The evaluation shows that GPT-5.2 performs best with a workflow that detects entity types for each input table individually, while Qwen-3.6-27B achieves better results on a workflow where the entity types are detected on an integrated table constructed from all input tables. The models achieve a Completeness rate of at least 97\% in entity type detection, at least 81\% for integrated attribute detection,
    and an 84\% F1-score for mapping the input columns to the correct integrated attributes.
    
    \item We perform an ablation study to prove the utility of the self-consistency strategy and the inclusion of a review loop in the Schema Matching Operator. The review loop increases the F1-score of the mappings by at least 10\% for both models, while the consistent operator outputs created using self-consistency increase the mapping F1-score for Qwen by 13\% and by 3\% for GPT-5.2 when compared to the averaged workflow results of three runs.

\end{enumerate}

\textbf{Structure of the paper:}
Sections~\ref{sec:operators} and \ref{sec:self-consistency} describe the SINT-Flow operators and how the self-consistency strategy is used on the operator outputs. Section \ref{sec:workflows} explains how the operators are arranged into different schema integration workflows. Section~\ref{sec:sint-benchmark} introduces the SINT-Bench schema integration benchmark, which we use to evaluate and compare the different workflows. Section~\ref{sec:experimental-setup} introduces the experimental setup, while Section~\ref{sec:experimental-results} presents the experimental results and an ablation study. Section~\ref{sec:baselines} compares the attribute clustering abilities of related work to that of the Schema Matching Operator. Section~\ref{sec:related-work} compares SINT-Flow to related work. Section~\ref{sec:conclusion} summarizes our conclusions. Section~\ref{sec:artifacts} links to the SINT-Flow repository, which contains all code, data, and ground truth files for replicating the experiments.

\section{The SINT-Flow Framework}
\label{sec:operators}
This section describes the SINT-Flow framework and its operators. The framework consists of five operators that can be arranged in different orders into schema integration workflows.

\begin{figure}
  \centering
  \includegraphics[width=\columnwidth]{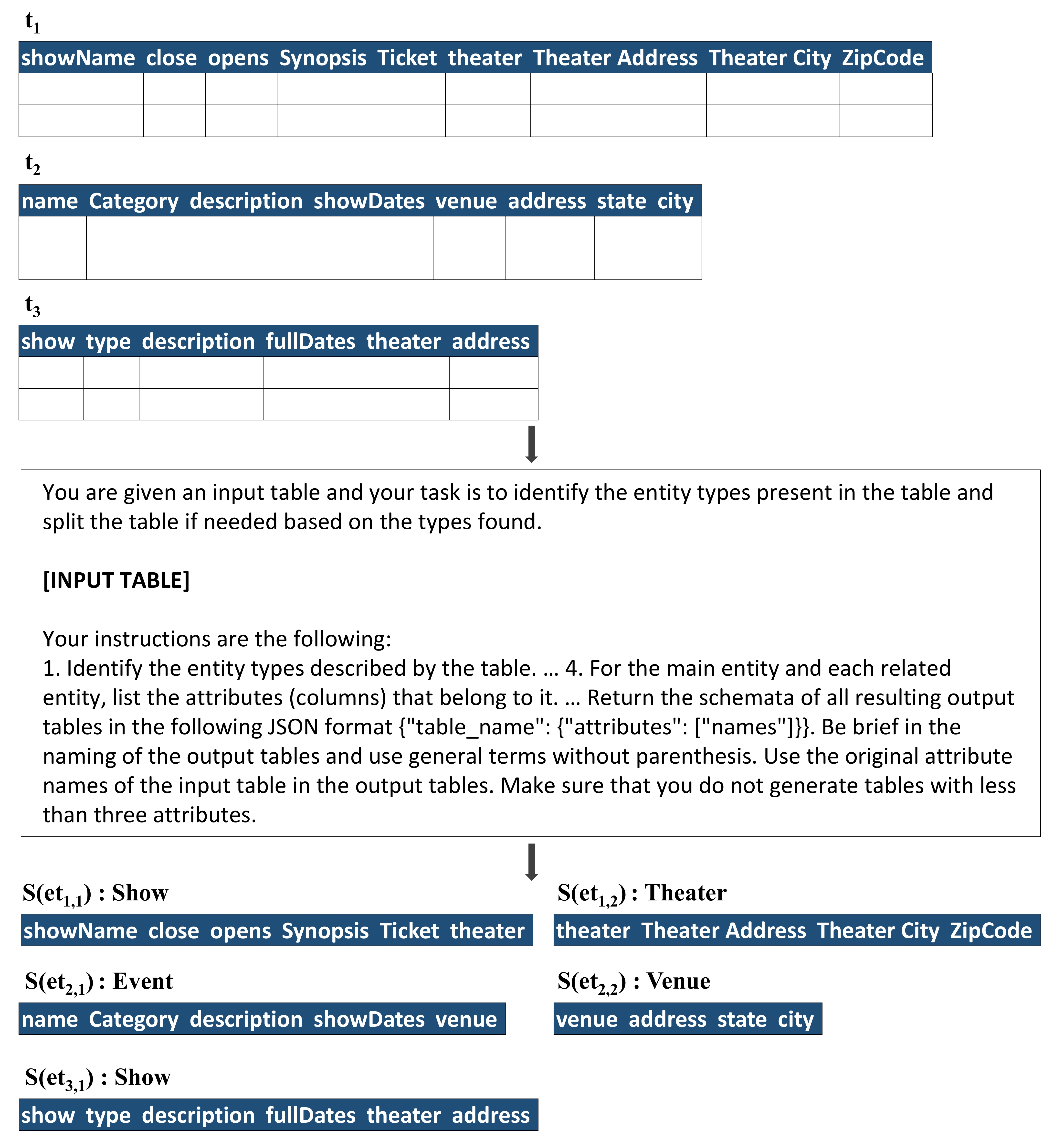}
  \caption{Table Splitting Operator: an example.}
  \label{fig:table-splitting-operator}
\end{figure}

\subsection{Table Splitting Operator} 
\label{subsec:table-splitting-operator}
The goal of this operator is to detect the relevant entity types in a given input table and to assign each table column to its associated entity type. We consider an entity type to be relevant if at least three columns can be associated with it; otherwise, the operator does not include that entity type in its response. The reason why we employ this minimum number is to avoid splitting an input table into overly narrow output tables.

\textbf{Input/Output.} The input to the operator is a set of tables $T=\{t_1, t_2, ..., t_n\}$ where $S_T = \{S_{t_1}, S_{t_2}, \dots, S_{t_n}\}$ is the set of schemata (i.e. collection of columns) of the tables $T$ and the output is the set of entity type table schemata $S_{T_E} = \{S_{et_{1,1}}, S_{et_{1,2}},\dots, S_{et_{n,m}}\}$, where $S_{et_{n,m}}$ is the collection of columns assigned to the $m^{th}$ entity type detected in table $t_n$. All columns of the input tables are assigned to an entity type table schema in $S_{T_E}$.

\textbf{Response Verification.} The operator provides the LLM with a tool to check the number of columns per detected entity type, so that the minimum number can be verified, ensuring that the LLM has followed the rule included in the prompt. We also introduce a post-splitting check to verify that all table columns have been assigned to at least one of the entity types detected. In the case where columns are missed, an LLM is asked to assign the missed columns to one of the detected types.

\textbf{Running Example.} Figure~\ref{fig:table-splitting-operator} shows an example of the input, prompt, and output of the Table Splitting Operator. The upper part of the figure shows the set of input tables $T = \{t_1, t_2, t_3\}$. The three tables describe events/shows and the venues where they are taking place. As shown in the bottom part of the figure, the Table Splitting Operator has generated the schema for entity type table \textit{Show} ($S_{et_{1,1}}$) and \textit{Theater} ($S_{et_{1,2}}$) derived from $t_1$ and \textit{Event} ($S_{et_{2,1}}$) and \textit{Venue} ($S_{et_{2,2}}$) derived from $t_2$ by assigning to them the columns that are associated with each type. In the last input table $t_3$, a \textit{Theater} entity type could also be detected. However, since the columns that can be assigned to \textit{Theater} are only two (\textit{theater}, \textit{address}), only the schema for the entity type table \textit{Show} ($S_{et_{3,1}}$) is generated from this table.

\textbf{Entity Tables versus Full Normalization.} Our framework does not aim to fully normalize the input tables. Instead, it focuses on creating entity-specific integrated schemata in which each entity type is described by at least three attributes. The minimum number of attributes is motivated by the use-case requirement to avoid overly narrow tables. We target use cases where no documentation or data dictionaries are available. This requires the discovery of functional dependencies (FDs) to be purely data-driven, which is known to be error-prone~\cite{berti2018discovery,kossmann2022data}. 
To illustrate the problems that arise from discovering FDs in our setup, we conduct an experiment by sampling 50 tables from SINT-Bench and comparing manually discovered FDs to FDs discovered using the TANE algorithm~\cite{huhtala1999tane} implemented in the \textit{desbordante}\footnote{\url{https://github.com/Desbordante/desbordante-core}} library. The experiment resulted in 632 manually discovered FDs compared to 3,316 FDs discovered by TANE. By analyzing the TANE-discovered FDs, we conclude that while the FDs hold given the data, most of them are not meaningful given knowledge about the application domain. This is a result of the limited domain coverage of the tables, making TANE discover FDs that overfit to misleading patterns in the data.

\begin{figure}
  \centering
  \includegraphics[width=\columnwidth]{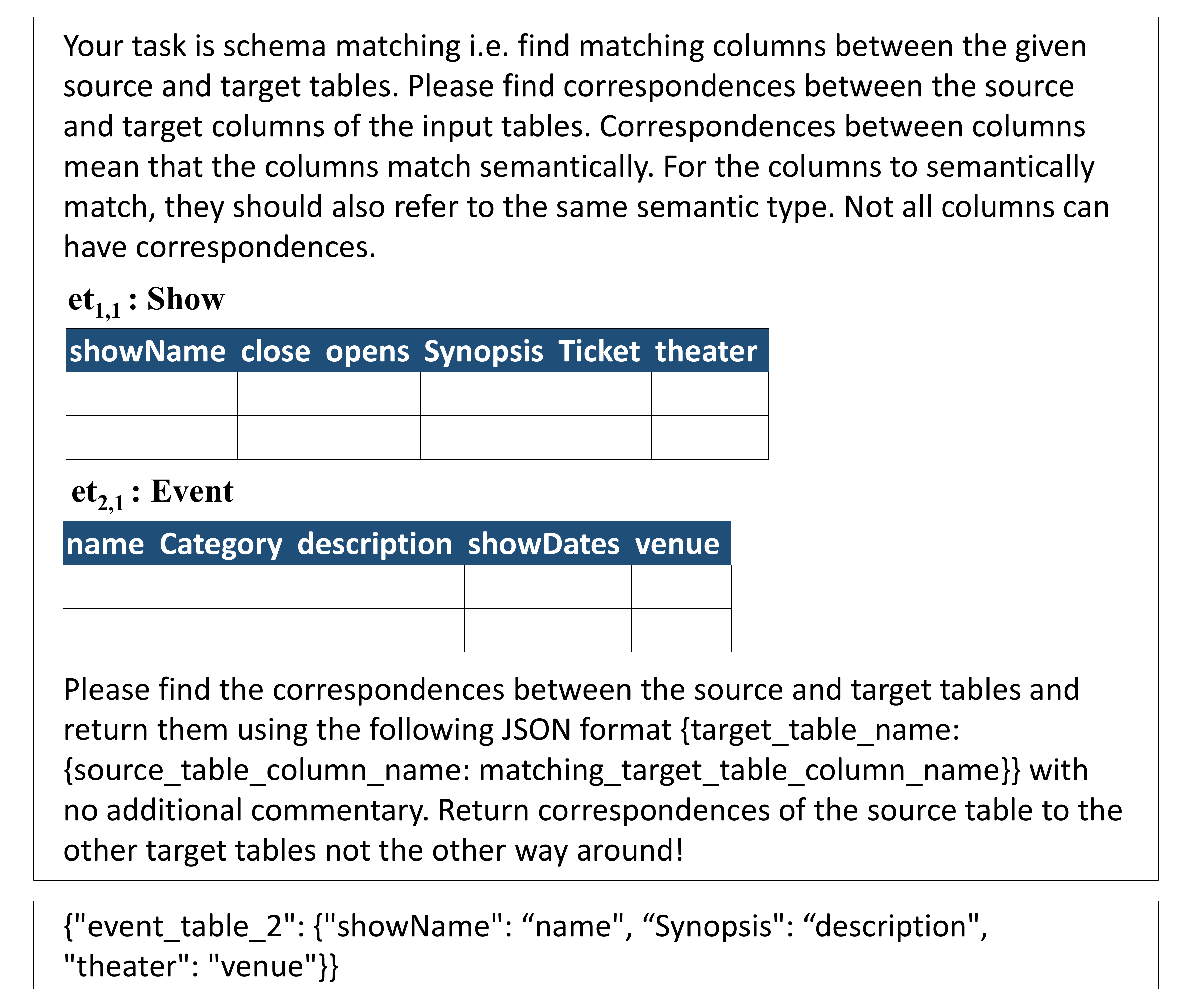}
  \caption{Schema Matching Prompt: an example.}
  \label{fig:schema-matching-prompt}
\end{figure}

\subsection{Schema Matching Operator} Schema matching is the task of discovering correspondences (matches) between attributes in different schemata~\cite{rahm2001survey}. The Schema Matching Operator's task is to find matching pairs of columns in the input tables that refer to the same semantic concept. Based on previous work on schema matching for tabular data~\cite{li2024tablegpt}, we prompt the LLM to find the matching columns in a table-pair-wise manner, i.e., we pass the input tables in pairs and ask the LLM to return all correspondences between the columns of the tables. Figure~\ref{fig:schema-matching-prompt} shows an example of a schema matching prompt together with a response from the LLM. 
The matches found between the columns are used to create attribute groups where each attribute group includes columns that were matched directly or indirectly, e.g. if column $zipcode$ from table \textit{a} is matched to $code$ from table \textit{b} and $zip$ from table \textit{c} is matched to $zipcode$ from table \textit{a}, one attribute group formed would be $(table_a.zipcode, table_b.code, table_c.zip)$. 

\textbf{Input/Output.} The input to this operator is pairs of entity type tables 
from $T_E$ (or $T$ depending on the sequence of the operators) and the output is a set of correspondences between their columns $Corr = \{(et_{1,1}.c_1, et_{2,1}.c_1), \dots, (et_a.c_x, et_b.c_y)\}$ and the formed attribute groups $AG$ based on the discovered correspondences $Corr$.

\textbf{Post-Schema Matching Reviewer Loop.} Because column correspondences form the basis for constructing attribute groups, false-positive matches can substantially compromise the resulting groups. A wrong correspondence between columns would lead to the merging of two attribute groups that should be separate and a loss of an integrated attribute. For this reason, we introduce a post-schema matching LLM reviewer loop to review the matches while forming the attribute groups. The idea behind this loop is to check whether a column A that is matched to a column B can also be matched to at least one other column to which column B is matched. We start forming the attribute groups by looping over the column correspondences and retrieve for each column the attribute group they are currently assigned to. At the beginning of the process, the set of groups $AG$ is empty; therefore, in the first case when both columns do not belong to any group, the LLM reviewer is called to re-check if the match between these columns is correct and, if TRUE, both columns form a new group which is added to the overall set of groups. If only one of the columns is assigned to a group, we check if the \textit{unassigned} column was matched to any of the other columns in the group of the \textit{assigned} column, and if TRUE, the \textit{unassigned} column is added to this group; otherwise, the LLM reviewer is asked if the column can be added to this group. In the final case, if both columns have been assigned to separate groups, the LLM reviewer is directly called to review if these two groups can be merged. The post-schema matching loop returns the created attribute groups $AG$.

\subsection{Table Grouping Operator} 
This operator is used to group the entity tables derived from the output of the \textit{Table Splitting Operator} by entity type.

\textbf{Input/Output.} Formally, the input to this operator is a set of entity type tables $T_E = \{et_{1,1}, et_{1,2} ..., et_{n,m}\}$ that can be derived from the schemata $S_{T_E}$ generated by the \textit{Table Splitting Operator}. The output of the operator is the grouping of the tables under their respective entity type $E = \{E_1, E_2, ... E_k\}$ where $E_k$ is composed of all tables in $T_E$ that refer to the entity type $E_k$ and $k$ is the number of distinct entity types found in the input tables. At this operator, the previously created attribute groups $AG$ from the \textit{Schema Matching Operator} are split into attribute groups per entity type $AG_E$.

\textbf{Response Verification.} In this operator, we introduce several checks on the model response. First, we ensure that a table is assigned to only one entity type group. If a table is found in multiple groups, the table is deleted from all. Similarly, if a non-existing table is added, the table is deleted. As a next step, we prompt the LLM to add each missing table (including the tables deleted due to being in more than one group) to the post-processed groups.

\begin{figure}
  \centering
  \includegraphics[width=\columnwidth]{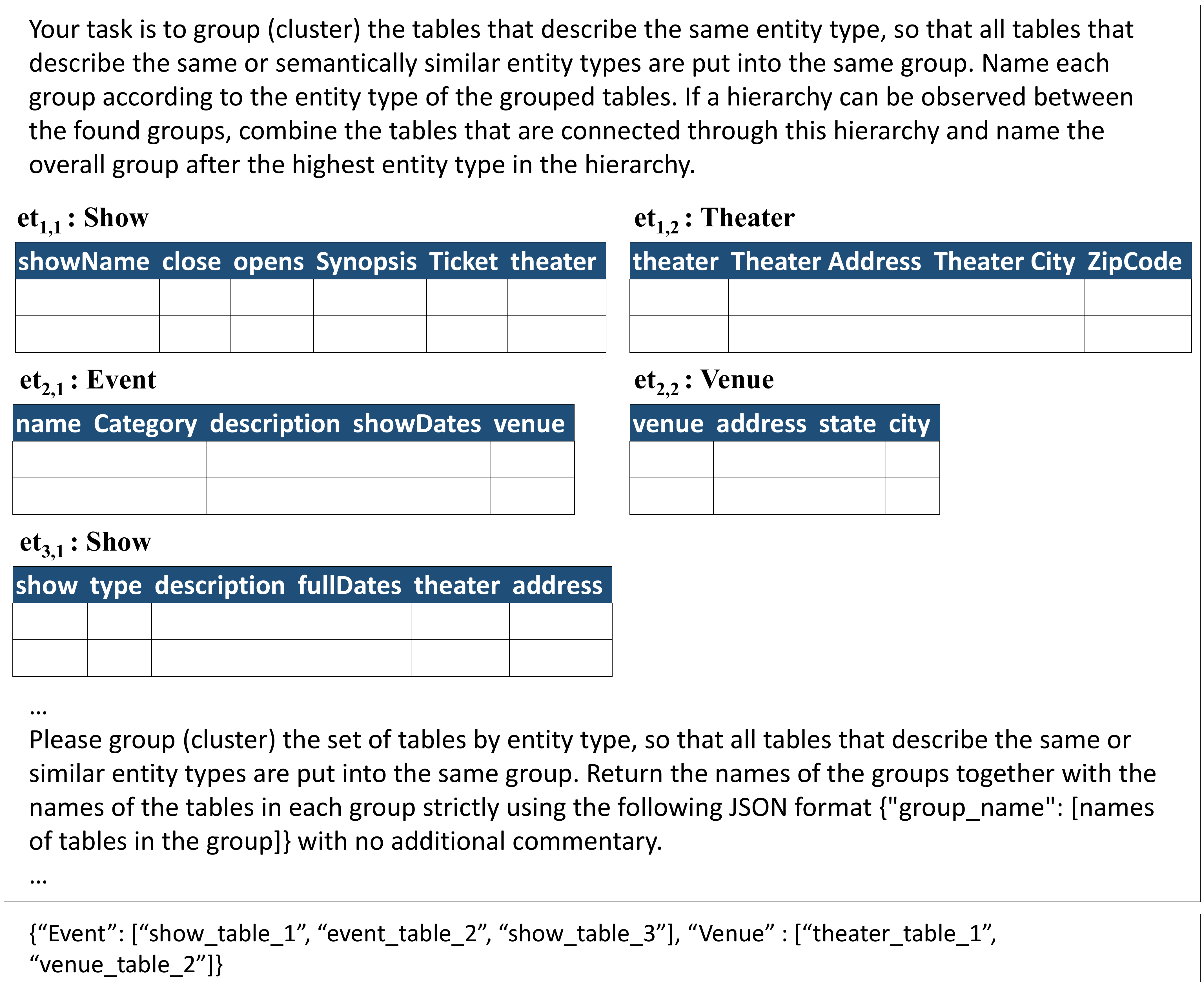}
  \caption{Table Grouping Prompt: an example.}
  \label{fig:table-grouping-operator}
\end{figure}

\textbf{Running Example.} Using the prompt that we show in Figure~\ref{fig:table-grouping-operator}, the output entity tables from the \textit{Table Splitting Operator} can be grouped as follows: entity tables $et_{1,1}$: \textit{Show}, $et_{2,1}$: \textit{Event} and $et_{3,1}$: \textit{Show} are grouped under the more general entity type \textit{Event}, while the entity tables $et_{1,2}$: \textit{Theater} and $et_{2,2}$: \textit{Venue} are grouped under the entity type \textit{Venue}. Overall, two distinct entity types have been detected in all input tables: $E= \{Event, Venue\}$.

\begin{figure*}
  \centering
  \includegraphics[width=\textwidth]{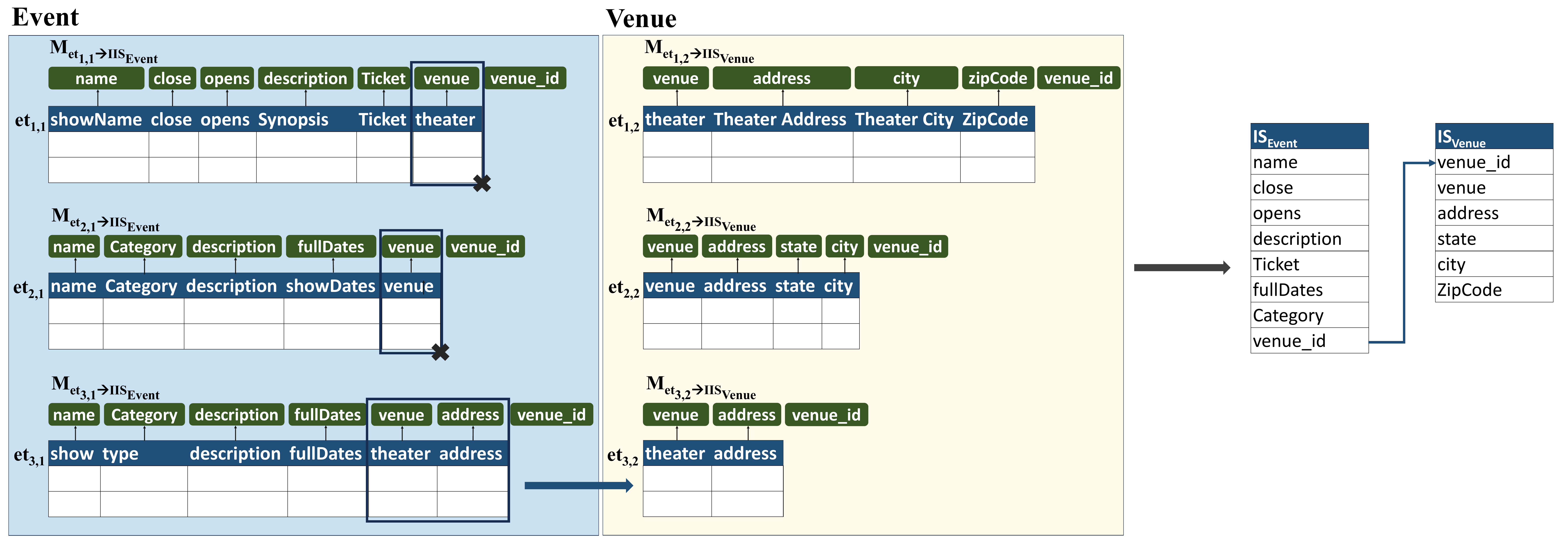}
  \caption{Integrated Schema Output Operator: example of its input and output.}
  \label{fig:final-output-operator}
\end{figure*}

\subsection{Attribute Merging Operator} The task of this operator is to choose for each attribute group a representative attribute from the group. The chosen attribute will be added to the intermediate integrated schema of its respective entity type and is called an \textit{integrated attribute}.

\textbf{Input/Output.} Specifically, the input to the operator is the attribute groups of each entity type $AG_E$, and the output of the operator is the intermediate integrated schema for each entity type $IIS_E$ and the mappings of all table columns in $S_{T_E}$ to an integrated attribute in the intermediate integrated schemata $M_{S_{T_E} \rightarrow IIS_{E}}$. In the case where the \textit{Table Splitting Operator} is not run before the \textit{Attribute Merging Operator}, $AG_E = AG$, i.e., the attribute groups have not been split by entity type yet. In the case where the \textit{Schema Matching Operator} is not used, i.e., attribute groups are not defined, the input to this operator is the entity tables $T_E$.

\textbf{Running Example.} An example of an attribute group for the Event type is ($et_{1,1}.showName$, $et_{2,1}.name$, $et_{3,1}.show$). The LLM in this case chooses \textit{name} as a representative attribute, as it can cover the values of all the columns in that group. By choosing an attribute for each group, the intermediate integrated schema of \textit{Event} can be built: $IIS_{Event}$ = $\{$name, close, opens, description, Ticket, venue, Category, fullDates, address$\}$. Along with the schema, the mappings of the input columns to the integrated attributes are output. For example, for the entity type table $et_{1,1}$, the mappings to the intermediate integrated schema of \textit{Event} are as follows: $M_{S_{et_{1,1}} \rightarrow IIS_{Event}}$ = \{``showName''$\colon$``name'', ``close''$\colon$``close'', ``opens$\colon$``opens'', ``Synopsis''$\colon$``description'',  ``Ticket''$\colon$``Ticket'', ``theater''$\colon$``venue''\}. Other mapping examples can be seen in the left part of Figure~\ref{fig:final-output-operator} above the columns of each entity table.

\subsection{Integrated Schema Output Operator} 
This operator has a fixed position at the end of the workflows as it should i) remove redundant attributes between the intermediate integrated schemata of entity types i.e. attributes that show up in more than one schema, ii) introduce identifiers if not already present, and iii) output foreign keys to link entity types.

\textbf{Input/Output.} Specifically, the input to the operator is the intermediate integrated schemata of all entity types $IIS_{E}$, and the output is the final set of integrated schemata $IS_{E}$ of all entity types, their foreign keys, and the mappings of all table columns to the integrated attributes in the integrated schemata $M_{S_{T_E}\rightarrow IS_{E}}$.

\textbf{Response Verification.} In this operator, we check whether the instruction to remove redundant attributes brings the complete removal of any attribute from all schemata. For this, we check that all integrated attributes present in the intermediate integrated schemata are assigned to at least one entity type schema. If this is not the case, we prompt the LLM to add the removed attributes to one of the schemata. 
This operator also provides a tool to the LLM that can check if any entity type schema is a subset of another so that redundant entity types can be removed.

\textbf{Running Example.} Figure~\ref{fig:final-output-operator} shows an example of the input and outputs of the operator. In the left part the entity type tables are mapped to the integrated attributes in the intermediate integrated schemata. First, the operator adds a \textit{venue\_id} identifier, which can be used to link the two entity types. Afterwards, it removes redundant attributes \textit{venue} and \textit{address} from the \textit{Event} schema, which are already present in the \textit{Venue} integrated schema. The removal of these two attributes brings the creation of a new entity table $et_{3,2}$ previously not created due to not meeting the minimum requirement in the \textit{Table Splitting Operator}. At this point, as it is known that the \textit{Venue} entity type can be found in other input tables and more than two attributes can be assigned to it, $et_{3,2}$ can be created. Finally, the mappings of the input columns are updated and the final integrated schemata for all entity types are output together with the mappings and foreign keys (shown in the right part of Figure~\ref{fig:final-output-operator}). The prompts used for all operators and the reviewer loop algorithm can be found in our GitHub repository linked in Section~\ref{sec:artifacts}.

\begin{figure}
  \centering
  \includegraphics[width=\linewidth]{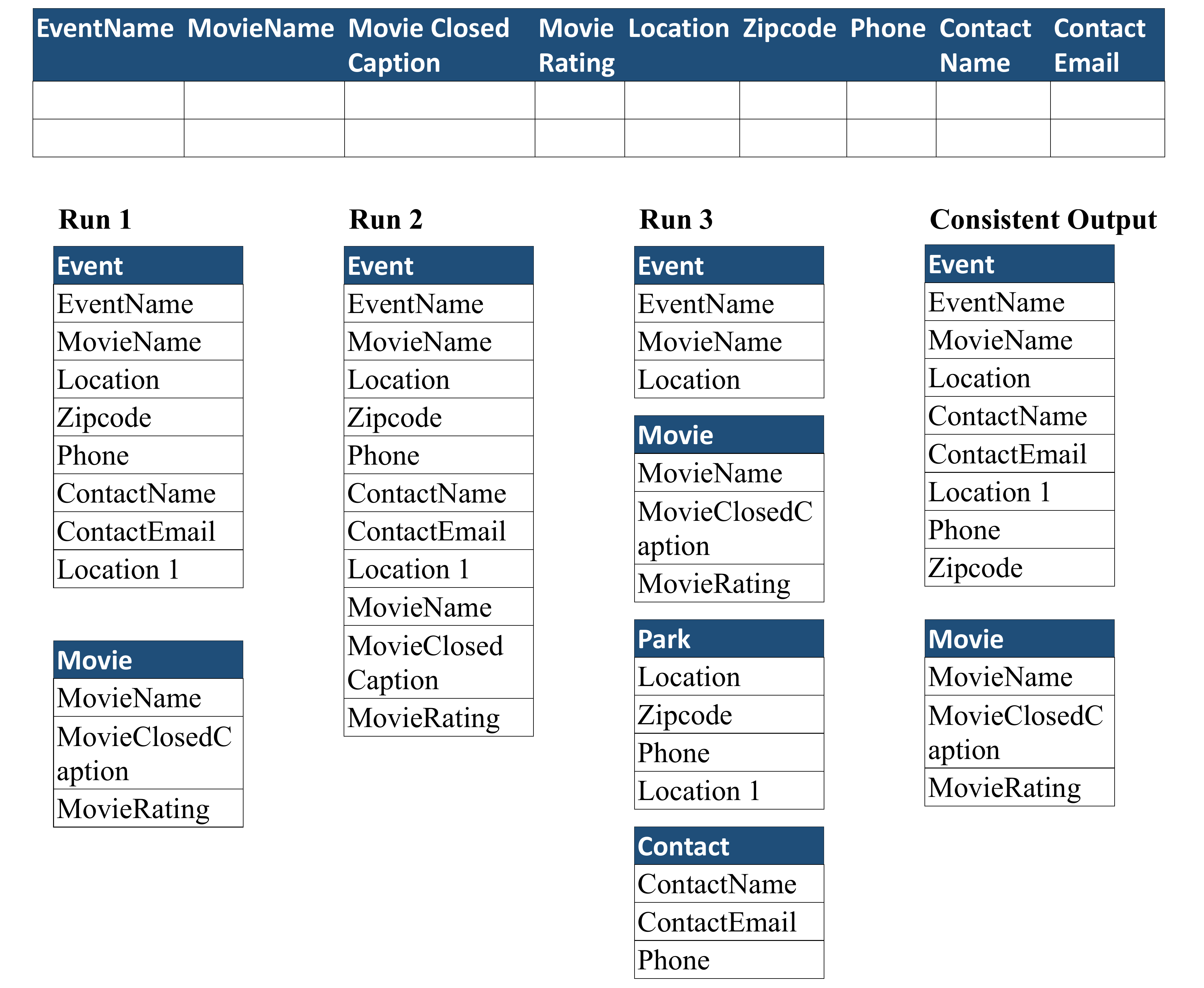}
  \caption{Table Splitting Operator: example of consistent output construction with self-consistency.}
  \label{fig:self-consistency}
\end{figure}

\section{Self-consistency for Operator Outputs}
\label{sec:self-consistency}

Self-consistency is a decoding strategy introduced by Wang et al.~\cite{wang2023selfconsistency} where an LLM is prompted multiple times with a high temperature in order to generate diverse reasoning paths that lead to an answer. The final response is constructed via majority voting and is considered to be a \textit{consistent} output. The idea behind the technique is that diverse reasoning paths should lead to the same result. Following this work, we integrate self-consistency by running three times each operator and the three responses are used to create \textit{consistent} outputs for each operator.

For the construction of consistent operator outputs, for some operators the majority voting method can be directly applied, while for others it can not. For the \textit{Schema Matching}, \textit{Attribute Merging} and the \textit{Integrated Schema Output} operators, majority voting can be applied directly. In \textit{Schema Matching}, if a match between two table columns is found in more than half the number of the runs, the correspondence is added to the \textit{consistent} output. In \textit{Attribute Merging}, the attribute that is chosen the most as a representative of each attribute group is added to the \textit{consistent} intermediate integrated schemata. The \textit{Integrated Schema Output Operator} consists of either removing redundant attributes, moving attributes between tables if they were misplaced, or adding foreign keys. For each addition and deletion of attributes and for each foreign key added, we check if this operation is done in more than half the runs, and, if true, this addition/deletion/foreign key is reflected in the \textit{consistent} output of the operator.

For the other two operators, \textit{Table Splitting} and \textit{Table Grouping}, majority voting can not be applied directly; therefore, we rely on co-occurrences between attributes/tables to construct the \textit{consistent} output for each operator, respectively. Our method counts, for each attribute/table, the number of times they occur in the same entity type/group with all other attributes/tables. After counting the co-occurrences, if the co-occurrence with an attribute/table is higher than half of the number of runs, we conclude that the pair of attributes/tables should be in the same entity type/group in the \textit{consistent} output. We loop through all attributes/tables and form the \textit{consistent} groupings of the attributes/tables for each operator, respectively.

\begin{table*}
    \caption{Statistics about the SINT-Bench tasks: the number of tables, average rows and columns, structure overlap, data types, average number of entity types per table, and the entity types per task with the number of their attributes in brackets.}
    \label{tab:schint}
    \centering
    \resizebox{\textwidth}{!}{
    \begin{tabular}{lccccccl}
    \toprule
        \textbf{Task Name} & \textbf{Tables} & \textbf{Avg. Rows} & \textbf{Avg. Cols} & \textbf{Structure Over.} & \textbf{Data Types} & \textbf{Entity Types} & \textbf{Entity Types (Integrated Attributes)} \\
        \midrule
        Volcanic Eruptions & 10 & 318 ($\pm220$) & 14 ($\pm2$) & 45\% ($\pm13$) & 59\% Numeric, 41\% Text & 1.6 ($\pm0.5$) & Volcano (15), Eruption (18) \\
        Movie Park Events & 6 & 235 ($\pm16$) & 9 ($\pm4$) & 56\% ($\pm6$) & 5\% Numeric, 88\% Text, 7\% Date & 2.0 ($\pm1.0$) & Movie (4), Park (6), Event (10) \\
        Events (GOBY) & 15 & 1016 ($\pm1993$) & 15 ($\pm3$) & 59\% ($\pm10$) & 10\% Numeric, 84\% Text, 6\% Date & 1.9 ($\pm0.3$) & Event (20), Venue (11) \\
        Flight Information & 10 & 1736 ($\pm1692$) & 13 ($\pm2$) & 47\% ($\pm6$) & 17\% Numeric, 79\% Text, \% Date & 2.0 ($\pm1.0$) & Flight (6), Passenger (7), Airport (15) \\
        Games & 6 & 4784 ($\pm7148$) & 13 ($\pm3$) & 63\% ($\pm12$) & 58\% Numeric, 41\% Text, 1\% Date & 1.0 ($\pm0.0$) & Game (18) \\
        Car Listings & 10 & 335 ($\pm76$) & 10 ($\pm1$) & 71\% ($\pm10$) & 43\% Numeric, 57\% Text & 1.0 ($\pm0.0$) & Car Listing (14) \\
        Movie Characters & 4 & 38 ($\pm2$) & 10 ($\pm2$) & 64\% ($\pm5$) & 85\% Text, 15\% Date & 2.8 ($\pm0.4$) & Work (6), Character (5), Actor (6) \\
        Paintings & 11 & 1208 ($\pm1245$) & 12 ($\pm3$) & 41\% ($\pm8$) & 23\% Numeric, 73\% Text, 4\% Date & 2.8 ($\pm0.5$) & Painting (13), Artist (10), Location (9) \\
        Publications & 13 & 155 ($\pm68$) & 13 ($\pm3$) & 58\% ($\pm3$) & 14\% Numeric, 77\% Text, 10\% Date & 1.6 ($\pm0.5$) & Publication (14), Journal (8) \\
        Monuments & 8 & 1260 ($\pm857$) & 9 ($\pm2$) & 44\% ($\pm14$) & 20\% Numeric, 80\% Text & 1.8 ($\pm0.4$) & Monument (14), Location (7) \\
        \bottomrule
    \end{tabular}}
\end{table*}

\textbf{Self-consistency Output Example.} We show an example of the Table Splitting self-consistency output construction in Figure~\ref{fig:self-consistency}. In the figure, a table containing events of movie screenings in parks is the input. At this step, the relevant entity types in the table should be detected and the correct attributes should be assigned to each detected type. In this table, there are three main entity types: \textit{Movie}, \textit{Park} and \textit{Event}. 
In the first run, the LLM has detected only two of the types; in the second run, it has detected only the \textit{Event} entity, and in the last, it has detected all three correct entities with the addition of a fourth entity \textit{Contact}. To form the self-consistency output, we check the co-occurrences of all attributes of the table. From the co-occurrences, we can determine that the attributes listed under the \textit{Contact} entity in the last run have been grouped with EventName and MovieName in a total of two runs; therefore, all these attributes should be assigned to the same entity type and the \textit{Contact} entity is not transferred to the consistent output. Similarly, the attributes in the \textit{Park} entity detected in the last run co-occur in two runs with the attributes of the \textit{Event} entity in the other runs; therefore, the \textit{Park} entity is not transferred to the consistent output and its attributes are put in the \textit{Event} entity. This leads to the construction of the \textit{consistent} output on the right side of the figure.

\textbf{Usage of LLMs for Schema Integration.} While LLMs have demonstrated strong performance across various domains, their outputs can differ across runs. Compared to traditional deterministic schema integration methods, this remains a weakness of methods using LLMs. To counter this answer variance, we apply the self-consistency strategy described above. We also conduct an ablation study on its usage in Section~\ref{subsubsec:self-consitency-res}.
LLMs are also known to hallucinate~\cite{huang2025survey}. This motivates the response verification loops in each operator as post-answer checks and we report their usage frequency in Section~\ref{subsec:per-operator-results}. Finally, the performance of LLMs degrades for very specialized domains that were unseen during training~\cite{shen2024tag}. In this work, we do not investigate methods for adapting LLMs to unseen domains and leave this as a direction for future work.

\section{The Schema Integration Workflows}
\label{sec:workflows}
Using the operators described in Section~\ref{sec:operators}, we build three different schema integration workflows. Each of the workflows applies the operators in a different order. 
All workflows take as input a set of tables and output integrated schemata for representing the data of the input tables. The workflows are described below:

\begin{enumerate}
    \item \textit{Entity Type Detection on Input Tables Workflow}: This workflow starts with the Table Splitting Operator, which detects the entity types for each input table separately. Afterwards, the Schema Matching Operator identifies matching columns between the entity tables, reviews the found matches and deletes likely false positives. Next, the Table Grouping Operator is used to group all tables that refer to the same entity type. Afterwards, the Attribute Merging Operator is applied to select the integrated attributes and build the intermediate integrated schema per entity type. Finally, the Integrated Schema Output Operator adds identifiers, removes redundant attributes and outputs the integrated schema per entity type and the mappings of the input columns to the integrated attributes.
    
    \item \textit{Implicit Schema Matching Workflow}: This workflow is similar to the first workflow but does not use the Schema Matching Operator. In this workflow, the tasks of the Schema Matching and Attribute Merging operators are combined. We do not ask the LLM to output the correspondences between the tables, but we input all entity tables and ask the LLM to output for each table the mappings of the columns to a chosen integrated attribute, where matching columns are mapped to the same integrated attribute. The motivation behind this workflow is to test the ability of the LLMs to process a large number of tables and be able to map each column to an integrated attribute without relying on an intermediate schema matching step, which creates attribute groups beforehand.
    
    \item \textit{Entity Type Detection on an Integrated Table Workflow}: In this workflow, we position the Table Splitting Operator at the end of the workflow. The workflow starts with the Schema Matching and Attribute Merging operators. The attribute groups and the intermediate integrated schema are not divided by entity type at this point of the workflow, in contrast to the other workflows. Afterwards, the Table Splitting Operator is applied to the single resulting \textit{integrated table}, whose columns are the integrated attributes in the intermediate schema and which is populated by transferring the entities (without deduplication) and their values (without value transformation) of all input tables into the corresponding columns using the mappings generated by the Attribute Merging Operator. 
    At the end, the Integrated Schema Output Operator is used. The motivation of applying the Table Splitting Operator at the end of the workflow is to test whether using a single integrated table makes entity type detection easier. The Table Grouping Operator is not required in this workflow as the table splitting is done on the integrated table.

\end{enumerate}

\section{The SINT-Bench Benchmark}
\label{sec:sint-benchmark}

This section introduces the SINT-Bench benchmark for evaluating schema integration methods. SINT-Bench consists of 10 schema integration tasks requiring the integration of altogether 93 different tables. Eight of the tasks involve 2 or more entity types and
require splitting input tables by
entity type. To the best of our knowledge, no other schema integration benchmark considers splitting input tables by entity type and assigning the table columns to the resulting types. We build SINT-Bench by collecting tabular data from existing benchmarks and defining for each input table a ground truth for the detection of entity types and the assignment of the table columns to these types. In addition to this table splitting ground truth, we also specify ground truth for the output of each operator: a table grouping ground truth that groups the split tables according to their entity types, a schema matching ground truth (in the use cases where it doesn't already exist) and an integrated schema for each entity type. The integrated schema includes entity identifiers as well as foreign key relationships connecting the schemata.

\textbf{Data Collection.} We gather the tabular data for SINT-Bench from various existing benchmarks. The data from the Events task has been sampled from the GOBY benchmark~\cite{kayali2024mind}, which includes real-world tabular data on different events and already provides a mapping of all columns to a common schema. The data from the Volcanic Eruptions, Flight Information, Games and Car Listings tasks are collected from the GDS benchmark~\cite{wu2025schema} which was built by collecting tabular data using Google Dataset Search~\cite{brickley2019google}. The GDS benchmark provides mappings of columns and tables to Schema.org types. As these tables had similar structures and column headers, we randomly dropped some columns from the tables and renamed the column headers using an LLM to make the integration use case more difficult. The Movie Park Events task is retrieved from the Join Benchmark~\cite{khatiwada2022integrating}, where the unified schema is already provided, thus we only divide this schema for the entity types detected in the tables. The remaining tasks were built using data from WikiDBs~\cite{vogel2024wikidbs}, which is a large corpus of 100,000 relational databases extracted from Wikidata~\cite{vrandevcic2014wikidata}. As the tables provided are already normalized, we create denormalized tables by joining on the foreign keys provided and randomly dropping up to four columns per new table created. To further increase heterogeneity, we rename column headers using an LLM and in the Paintings and Monuments tasks, we add some tables from other sources discovered using Google Dataset Search. For these tasks, we manually map the columns to integrated attributes and create their integrated schema.

\textbf{Statistics.} Table~\ref{tab:schint} gives an overview of the SINT-Bench tasks. The tasks require the integration of the schemata of 4 to 15 tables, each containing 9 to 15 columns. The GOBY task features the widest tables with an average of 15 columns. Two tasks, Games and Car Listings, include only 1 entity type per table, while the other 8 tasks cover 2 and 3 entity types. To find how similar the tables within the tasks are to each other, we calculate the \textit{structure overlap} by averaging the Jaccard similarity between tables of a task, calculated as the number of semantically matching columns between two tables over the total number of columns in the tables.
Higher structure overlap shows a higher similarity between the columns of the tables within a task; e.g., the tables of the Car Listings task have the highest number of overlapping columns.  
The data type mostly covered in the tables is textual values, and in some tasks numeric data is present in more than 40\% of the columns. The \textit{date} type is the least covered, with a maximum of 15\% date columns in the Movie Characters task.

\begin{table}
    \centering
    \caption{Statistics of related tabular benchmarks.}
    \label{tab:comparison-benchmarks}
    \resizebox{\columnwidth}{!}{
    \begin{tabular}{lccccc}
    \toprule
         \textbf{Name} & \textbf{Tasks} & \textbf{Tables} & \textbf{Width} & \textbf{Length} & \textbf{Structure} \\
         & & & & & \textbf{Overlap} \\
         \midrule
         Real Benchmark & 11 & 102 & 12 ($\pm$2) & 2166 ($\pm$851) & 67\% ($\pm$9)\\
         GDS & 42 & 660 & 23 ($\pm$7) & 1090 ($\pm$579) & 61\% ($\pm$11)\\
         WDC & 47 & 601 & 354 ($\pm$249) & 6 ($\pm$2) & 38\% ($\pm$9)\\ 
         BURR-Synthetic & 54 & 117 & 4 ($\pm$2) & 518 ($\pm$192) & -\\
         BURR-Real & 4 & 134 & 12 ($\pm$8) & 1130 ($\pm$1348) & -\\\midrule
         SINT-Bench & 10 & 93 & 12 ($\pm$2) & 1108 ($\pm$1332) & 55\% ($\pm$9)\\
         \bottomrule
    \end{tabular}}
\end{table}

\textbf{Comparison to related benchmarks.} Some existing tabular benchmarks for schema integration include the Real Benchmark~\cite{khatiwada2022integrating}, GDS~\cite{wu2025schema} and WDC benchmarks~\cite{wu2025schema}. The Real Benchmark includes tables divided into tasks, while in the latter ones the tables are not divided so that methods can be evaluated on the detection of table topics. The distinction of SINT-Bench from these existing benchmarks is the inclusion of the table splitting ground truth. For each input table, the columns in the table are grouped according to entity types. For each entity type, SINT-Bench provides a ground truth for its integrated schema, which includes foreign keys and added identifiers. Closely related is also the benchmark for ontology learning from relational databases, BURR~\cite{laskowski2025burr}. The benchmark consists of 54 LLM-generated ontology learning tasks, organized around aspects such as normalization degree and relationship cardinality, and 4 real-world use cases 
with databases containing between 8 and 71 relational tables. Table~\ref{tab:comparison-benchmarks} compares SINT-Bench with the other benchmarks.

\section{Experimental Setup}
\label{sec:experimental-setup}

The following sections present the evaluation of SINT-Flow using the SINT-Bench benchmark. We evaluate the final output of the workflows, as well as the results of each operator. This section provides details on the employed LLMs, the table serialization format, and the evaluation metrics used in the experiments. All code and data for replicating the experiments are available in the GitHub repository referenced in Section \ref{sec:artifacts}.

\textbf{Models.} We evaluate the workflow using the hosted LLM GPT-5.2\footnote{\url{https://developers.openai.com/api/docs/models/gpt-5.2}} (full version name: gpt-5.2-2025-12-11) and the open-weight model Qwen-3.6-27B\footnote{\url{https://huggingface.co/Qwen/Qwen3.6-27B}} as backbone models. We test both types of models to also cover use cases where hosted LLMs can not be employed due to privacy concerns. We prompt and invoke GPT-5.2 using the LangChain library\footnote{\url{https://www.langchain.com/}} and Qwen-3.6-27B using the HuggingFace library\footnote{\url{https://huggingface.co/}}. For Qwen, we use the recommended settings for non-thinking mode: temperature=0.7, top\_p=0.80, top\_k=20, while for GPT-5.2 we use temperature=1.

\textbf{Table Serialization.} We serialize the tables in Markdown format, including their column headers. Before serialization, we sort the rows of each table by the number of non-empty values in descending order, ensuring that the most complete rows appear first. We then include the first 10 rows of each table in the prompt.

\textbf{Evaluation Metrics.} We evaluate two artifacts generated by SINT-Flow: the generated \textit{Integrated Schema} and the \textit{Mappings} of all source columns to the integrated attributes. We employ two main metrics to measure the quality of the generated \textit{Integrated Schema} with regard to the criteria set by Batini et al.~\cite{batini1986comparative}: \textbf{1) Completeness}, which reflects how many concepts in the source schemata can be modeled by the generated integrated schema~\cite{batista2007information}. We calculate completeness as $C = |S_{pred} \cap S_{gt}| / |S_{gt}|$, where $|S_{gt}|$ is the number of elements in the ground truth schemata and $|S_{pred}|$ is the number of predicted elements in the generated schemata. \textbf{2) Minimality}, which reflects the redundancy present in the generated schema~\cite{batista2007information,bellahsene2010evaluating}. The higher the minimality, the less redundant the generated schema is. We calculate minimality as $M = 1- ( ( |S_{pred}| - |S_{pred} \cap S_{gt}| ) / |S_{gt}| )$. We report the completeness and minimality of both the \textit{Entity Types} and \textit{Attributes} in the schema. For the \textit{Mapping} between input columns and integrated attributes as well as for all other operator results, we report the Precision, Recall, and F1-score. To calculate all metrics, the predicted entity types are first mapped to the entity types in the ground truth based on the overlap between attributes assigned to the types. Similarly, the predicted attributes are mapped to the ground truth attributes using the overlap between column values of the columns mapped to these attributes.

\section{End-to-End Evaluation}
\label{sec:experimental-results}
This section presents the evaluation results of the workflows, compares them to single-prompt baselines, and conducts a scalability analysis and ablation study.

\begin{table}
    \centering
    \caption{Results per workflow: minimality (M) and completeness (C) of the entity type and the attribute detection, precision (P), recall (R) and F1 of the mapping generation.}
    \label{tab:workflow-setups}
    \resizebox{\columnwidth}{!}{
    \begin{tabular}{lcccccccc}
         \toprule
         & \textbf{Work-} & \multicolumn{2}{c}{\textbf{Entities}} & \multicolumn{2}{c}{\textbf{Attributes}} & \multicolumn{3}{c}{\textbf{Mappings}}\\
         \textbf{Model} & \textbf{flow} & \multicolumn{1}{c}{\textbf{M}} & \multicolumn{1}{c}{\textbf{C}} & \multicolumn{1}{c}{\textbf{M}} & \multicolumn{1}{c}{\textbf{C}} & \textbf{P} & \textbf{R} & \textbf{F1}\\
         \midrule
         \multirow{3}{*}{\textbf{Qwen-27B}} & 1 & 0.80 & 0.93 & 0.76 & 0.80 & 0.72 & 0.70 & 0.71\\
         & 2 & 0.80 & \textbf{1.00} & 0.54 & \textbf{0.88} & 0.80 & 0.73 & 0.76\\
         & 3 & \textbf{0.97} & 0.97 & \textbf{0.90} & 0.81 & \textbf{0.87} & \textbf{0.81} & \textbf{0.84}\\
         \midrule
         \multirow{3}{*}{\textbf{GPT-5.2}} 
         & 1 & 0.88 & \textbf{1.00} & \textbf{0.85} & 0.87 & \textbf{0.86} & \textbf{0.83} & \textbf{0.84}\\
         & 2 & 0.83 & \textbf{1.00} & 0.55 & \textbf{0.92} & 0.79 & 0.74 & 0.76\\
         & 3 & \textbf{1.00} & 0.88 & 0.85 & 0.81 & 0.83 & 0.80 & 0.81\\
         \bottomrule
    \end{tabular}}
\end{table}

\begin{table*}
    \centering
    \caption{Detailed operator results for the three LLM workflows per model.}
    \label{tab:workflow-operators}
    \resizebox{\textwidth}{!}{
    \begin{tabular}{lccccccccccccccccc}
         \toprule
         \textbf{Model} & \textbf{Workflow} &\multicolumn{4}{c}{\textbf{Table Splitting}} & \multicolumn{3}{c}{\textbf{Grouping}} & \multicolumn{3}{c}{\textbf{Schema Matching}} & \multicolumn{3}{c}{\textbf{Attribute Merging}} & \multicolumn{3}{c}{\textbf{Integrated Schema}} \\
         
         & &\multicolumn{2}{c}{\textbf{Entity Types}} & \multicolumn{2}{c}{\textbf{Attributes}} &  &  &  &  & & & \multicolumn{3}{c}{\textbf{Mappings}} & \multicolumn{3}{c}{\textbf{Mappings}} \\
         
         & & \textbf{M} & \textbf{C} & \textbf{M} & \textbf{C} & \textbf{P} & \textbf{R} & \textbf{F1} & \textbf{P} & \textbf{R} & \textbf{F1} & \textbf{P} & \textbf{R} & \textbf{F1} & \textbf{P} & \textbf{R} & \textbf{F1} \\
         \midrule
        \multirow{3}{*}{\textbf{Qwen-27B}} 
        & 1 & 0.95 & 0.87 & 0.91 & 0.87 & 0.88 & 0.82 & 0.84 & 0.94 & 0.94 & 0.94 & 0.78 & 0.75 & 0.77 & 0.72 & 0.70 & 0.71 \\
        & 2 & 0.89 & 0.89 & 0.87 & 0.83 & 0.88 & 0.85 & 0.86 & - & - & - & 0.78 & 0.75 & 0.76 & 0.80 & 0.73 & 0.76 \\
        & 3 & 0.79 & 0.98 & 0.87 & 0.83 & - & - & - & 0.96 & 0.95 & 0.95 & 0.93 & 0.93 & 0.93 & 0.87 & 0.81 & 0.84 \\
         \midrule
        \multirow{3}{*}{\textbf{GPT-5.2}} 
        & 1 & 0.99 & 0.90 & 0.93 & 0.91 & 0.93 & 0.86 & 0.89 & 0.95 & 0.95 & 0.95 & 0.86 & 0.84 & 0.85 & 0.86 & 0.83 & 0.84 \\
        & 2 & 0.99 & 0.92 & 0.94 & 0.91 & 0.90 & 0.84 & 0.87 & - & - & - & 0.80 & 0.78 & 0.79 & 0.79 & 0.74 & 0.76 \\
        & 3 & 0.91 & 0.91 & 0.90 & 0.86 & - & - & - & 0.97 & 0.96 & 0.96 & 0.94 & 0.94 & 0.94 & 0.83 & 0.80 & 0.81 \\
         \bottomrule
    \end{tabular}}
\end{table*}

\subsection{Workflows' Results}
\label{subsec:workflow-results}

In Table~\ref{tab:workflow-setups}, we report the averaged metrics over the 10 tasks of SINT-Bench for the integrated schema and mappings output by the three LLM workflows. For brevity, the workflows in the table will be referred to by their assigned numbers in Section~\ref{sec:workflows}. The workflow that achieves the highest metric for all aspects is different for both models. For GPT-5.2 detecting the entity types individually for each input table (Workflow 1) leads to an average Completeness of 100\% for entity type detection and 87\% for assigning the correct integrated attributes and adding the correct identifiers to each entity type. For Qwen-3.6-27B the workflow where the entity types are detected on the \textit{integrated table} created from all input tables (Workflow 3) reaches an average Completeness of 97\% for entity type detection and 81\% for attribute detection. All these values translate into an overall mapping F1-score of 84\% for both GPT-5.2 and Qwen-3.6-27B. 

Regarding Workflow 3, we can see that GPT-5.2 is more reluctant to split the \textit{integrated table} according to entity types, which leads to a lower Entity Completeness rate, the lowest between all workflows, and affects the low Attribute Completeness, while for Qwen-3.6-27B this unified view of the input tables helps distinguish the entity types better, the Entity Minimality and Completeness stand both at 97\%. For both models, the implicit schema matching workflow (Workflow 2) produces the schema with the highest \textit{Completeness} rate compared to the other two workflows. Since this workflow does not use the Schema Matching Operator, the model misses matches between columns, which leads to the existence of more integrated attributes in the integrated schema. This is reflected in the Minimality, as Workflow 2 has the lowest Attributes Minimality, i.e., redundant attributes exist in the generated schema. This shows the advantage of finding the correspondences in a separate step and then creating the integrated schema, rather than combining both tasks in one step.

\subsection{Operators' Evaluation and Error Analysis}
\label{subsec:per-operator-results}
In this section, we present the detailed results of each operator used in the three LLM workflows in Table~\ref{tab:workflow-operators}. In this table, we do not repeat the entity type and attribute detection results for the last operator and refer to them in Table~\ref{tab:workflow-setups}. As the operators are run in sequence, the results of one operator affect the results of the subsequent operators, i.e., errors are propagated through the workflow. The Schema Matching Operator is an exception to this as its input is not dependent on any other operator's output.

\textit{Table Splitting Operator.} When detecting entity types using the Table Splitting Operator, GPT-5.2 is more reluctant to detect multiple entity types from the integrated table (Workflow 3) than on each table individually (Workflows 1 and 2) and we can see this from the 12\% lower Entity Completeness rate. This is in contrast to Qwen, which has a smaller Entity Completeness gap between the three workflows. By analyzing the errors made by GPT-5.2 for Workflow 1 for the attribute detection aspect, we find that 69\% of the errors are made because some entity types are not detected by the operator. As for the rest of the errors, 6\% of them are due to models not assigning overlapping columns to multiple entity types (e.g., the venue column should be assigned to both Event and Venue types) and 25\% of the errors are made because of wrong assignment of columns to the wrong entity types, due to semantically ambiguous attribute names.

\textit{Table Grouping Operator.} In the grouping operator, by conducting an error analysis on the GPT-5.2 results, we discover that only 35\% of the errors are made because of the ``grouping'' ability of the LLM operator. These errors are made due to the LLM not grouping entity types that belong to the same concept but are at different levels of the hierarchy (e.g., Show and Event). The remaining 65\% of the errors are influenced directly by the Table Splitting Operator, where most entity types are not detected and can therefore not be grouped.

\textit{Integrated Schema Output Operator.} In the Integrated Schema Mappings, we similarly see that 71\% of the errors are inherited from the other operators of the workflows, mainly the Table Splitting and Schema Matching operators. The other 29\% of the errors are made by the operator itself, where errors are made due to not adding identifiers when they are not present in the schema, and not removing redundant attributes.

\textit{Verification Loops.} We notice that GPT-5.2 uses the verification loop of the Integrated Schema Output Operator 4 times out of the overall 30 runs of the operator (10 tasks, 3 runs), while the verification loops of the Table Splitting and Integrated Schema Output operators are used 18 and 11 times by Qwen-3.6-27B, i.e. in 18 instances Qwen had forgotten to assign a column from an input table to at least one entity type detected and had entirely deleted 11 times an attribute from the integrated schemata. Neither model uses the verification loop of the Table Grouping Operator.

\begin{table}
    \centering
    \caption{Single-prompt baseline results: minimality (M) and completeness (C) of entity type and attribute detection, precision (P), recall (R) and F1 of the mapping generation.}
    \label{tab:single-prompt}
    \resizebox{\columnwidth}{!}{
   
    \begin{tabular}{llccccccc}
         \toprule
         & & \multicolumn{2}{c}{\textbf{Entities}} & \multicolumn{2}{c}{\textbf{Attributes}} & \multicolumn{3}{c}{\textbf{Mappings}}\\
         \textbf{Model} & \textbf{} & \textbf{M} & \textbf{C} & \textbf{M} & \textbf{C} & \textbf{P} & \textbf{R} & \textbf{F1}\\
         \midrule
         \multirow{2}{*}{\textbf{Qwen-27B}} & \textbf{CoT}    & 0.77 & 0.85 & 0.71 & 0.72 & 0.76 & 0.69 & 0.72 \\
         & \textbf{Instr.} & 0.58 & 0.88 & 0.81 & 0.71 & 0.79 & 0.68 & 0.72 \\
         \midrule
         \multirow{2}{*}{\textbf{GPT-5.2}}  & \textbf{CoT}    & 0.74 & 0.86 & 0.54 & 0.78 & 0.77 & 0.68 & 0.72 \\
          & \textbf{Instr.} & 0.81 & 0.89 & 0.64 & 0.79 & 0.77 & 0.71 & 0.74 \\
         \bottomrule
    \end{tabular}}
\end{table}

\begin{table*}
    \centering
    \caption{Average runtime (in minutes and seconds) and token usage per task for the different workflows.}
    \label{tab:token-time-analysis}
    \resizebox{\textwidth}{!}{
    \begin{tabular}{lccccccccccccc}
         \toprule
         \multirow{2}{*}{\textbf{Model}} & \multirow{2}{*}{\textbf{Workflow}} & \multicolumn{2}{c}{\textbf{Total}} & \multicolumn{2}{c}{\textbf{Table Splitting}} & \multicolumn{2}{c}{\textbf{Schema Matching}} & \multicolumn{2}{c}{\textbf{Table Grouping}} & \multicolumn{2}{c}{\textbf{Attribute Merging}} & \multicolumn{2}{c}{\textbf{Schema Output}} \\
         & & \textbf{Tokens} & \textbf{Time} & \textbf{Tokens} & \textbf{Time} & \textbf{Tokens} & \textbf{Time} & \textbf{Tokens} & \textbf{Time} & \textbf{Tokens} & \textbf{Time} & \textbf{Tokens} & \textbf{Time} \\
         \midrule
         \multirow{3}{*}{\textbf{GPT-5.2}} & 1 & 683,865 & 13:28 & 66,044 & 02:13 & 568,819 & 10:28 & 15,906 & 00:11 & 19,098 & 00:20 & 13,999 & 00:14 \\
         & 2 & 119,694 & 04:59 & 66,130 & 02:06 & - & - & 15,883 & 00:08 & 22,116 & 00:52 & 15,564 & 01:51\\
         & 3 & 279,333 & 05:05 & 6,411 & 00:18 & 249,481 & 04:29 & - & - & 16,243 & 00:10 & 7,198 & 00:06  \\
         \midrule
         \multirow{3}{*}{\textbf{Qwen-27B}} & 1 & 657,935 & 52:49 & 48,947 & 27:13 & 560,157 & 21:43 & 16,163 & 00:32 & 18,366 & 01:00 & 14,303 & 02:18 \\
         & 2 & 99,651 & 40:32 & 48,359 & 25:47 & - & - & 14,729 & 00:33 & 20,647 & 05:15 & 15,916 & 08:56\\
         & 3 & 312,010 & 23:58 & 6,241 & 04:16 & 278,810 & 14:39 & - & - & 16,459 & 00:47 & 10,590 & 04:15 \\
         \bottomrule
    \end{tabular}}
\end{table*}

\subsection{{Comparison to Single-prompt Baselines}}
\label{subsec:single-prompt-baseline}
In order to evaluate the need for using separate operators for the schema integration task, we establish two single-prompt baselines in which the model is prompted to create an integrated schema and to output the mappings of all input tables to the integrated attributes in a single model call. In this setting, all input tables are included in the single prompt. We establish two prompts: i) a \textit{Chain-of-Thought} (CoT) prompt where we task the model to generate the integrated schema from the input tables by planning its steps and outputting the schema, mappings and its reasoning, and ii) the \textit{Instruction} prompt where we provide detailed instructions for the steps to be followed to generate the integrated schema. Each step corresponds to an operator's tasks and we ask the model to output the intermediate results of all steps in addition to the integrated schema and the mappings. The results of both prompts are shown in Table~\ref{tab:single-prompt}. For both models, we observe that providing detailed instructions, compared to letting the model plan its steps in a CoT approach, changes the F1-score of the mappings minimally, but the Attribute Minimality is higher, i.e., fewer redundant attributes are present in the generated schema. We observe a severe drop in the Mappings F1-score (approx. minus 10 percentage points) between the results of the single-prompt baselines and those of the best-performing workflows (Table~\ref{tab:workflow-setups}). We conclude from this comparison that the decomposition of the schema integration task into multiple operators benefits the generation of the integrated schema and its mappings. Both prompts and their results are provided in the GitHub repository referenced in Section \ref{sec:artifacts}.

\begin{table}
    \centering
    \caption{Scalability experiments on sets of 25, 50, 75 and 100 tables sampled from the GOBY benchmark. The time is expressed in hours and minutes.}
    \label{tab:scalability}
    \resizebox{\columnwidth}{!}{
   
    \begin{tabular}{lrcccccccc}
         \toprule
         \multirow{2}{*}{\textbf{Model}} & \multirow{2}{*}{\textbf{}} & \multicolumn{2}{c}{\textbf{Entities}} & \multicolumn{2}{c}{\textbf{Attributes}} & \multicolumn{3}{c}{\textbf{Mappings}} & \multirow{2}{*}{\textbf{Time}} \\
         &  & \textbf{M} & \textbf{C} & \textbf{M} & \textbf{C} & \textbf{P} & \textbf{R} & \textbf{F1} & \\
         \midrule
         \multirow{4}{*}{\textbf{Qwen-27B}} 
        & 25 & 1.00 & 1.00 & 0.92 & 0.76 & 0.82 & 0.82 & 0.82 & 02:46 \\
        & 50 & 1.00 & 1.00 & 1.00 & 0.70 & 0.81 & 0.81 & 0.81 & 11:37 \\
        & 75 & 1.00 & 1.00 & 1.00 & 0.68 & 0.82 & 0.82 & 0.82 & 27:24 \\
        & 100 & 1.00 & 1.00 & 0.96 & 0.61 & 0.82 & 0.70 & 0.76 & 47:46\\
         \midrule
         \multirow{4}{*}{\textbf{GPT-5.2}}  & 25 & 1.00 & 1.00 & 0.76 & 0.76 & 0.83 & 0.83 & 0.83 & 00:19 \\
        & 50 & 1.00 & 1.00 & 0.85 & 0.70 & 0.79 & 0.79 & 0.79 & 01:25 \\
        & 75 & 1.00 & 1.00 & 0.86 & 0.68 & 0.79 & 0.79 & 0.79 & 04:00 \\
        & 100 & 1.00 & 1.00 & 0.79 & 0.61 & 0.75 & 0.63 & 0.69 & 07:08 \\
         \bottomrule
    \end{tabular}}
\end{table}

\subsection{{Scalability, Runtime and Token Usage}}
\label{subsec:token-time-analysis}
\textbf{Runtime and Token Usage.} Table~\ref{tab:token-time-analysis} shows the average runtime and token usage of the three workflows. The tokens reported include the input and output tokens and the tokens used for tool calls. We access GPT-5.2 using the OpenAI API, while we run Qwen-3.6-27B locally on a single NVIDIA RTX A6000 48GB GPU. Therefore, the reported runtime of the latter model is higher. Workflow 1 has the highest average runtime for both models, as the \textit{Table Splitting Operator} (TSO) has a runtime that is linear to the number of input tables, while the \textit{Schema Matching Operator} (SMO) has a quadratic runtime, due to all pairs of input tables being compared. SMO's average runtime is lower in Workflow 3, where the input tables have not been split into entity tables and TSO's runtime is lower as the detection of the entity types is done on the integrated table. Overall, Workflow 2 has the lowest average token usage and runtime; however, performance-wise, this setup reaches the lowest Mapping F1-score (see Table~\ref{tab:workflow-setups}).

\textbf{Scalability Analysis.} In order to further evaluate how the workflows perform as the number of input tables increases, we create four table sets of 25, 50, 75 and 100 tables sampled from the GOBY benchmark~\cite{kayali2024mind}, where we create a ground truth for table splitting. We report the results of running the workflow on these sets in Table~\ref{tab:scalability}. 
We observe that the F1-score of the mappings decreases as the number of input tables increases. An analysis of the model's errors reveals that this decline is primarily caused by the \textit{Schema Matching Operator} (SMO), particularly its post-matching review loop. As the number of columns per attribute cluster increases, the model becomes less likely to confirm previously identified correspondences, resulting in an increase in false negatives. These errors subsequently propagate to downstream operators, negatively affecting their results.
A further limitation of the current SMO implementation is its quadratic time complexity, which leads to increasing runtimes as the number of input tables grows. To address this limitation, we plan to develop a new workflow in which the \textit{Table Splitting Operator} (TSO) and the SMO operate in parallel. Rather than matching attributes across all input tables, the SMO would incrementally match each input table against an intermediate representation of the integrated schema. This representation would be updated after each matching step by adding attributes for which no correspondence is found. By avoiding pairwise comparisons between all input tables, this approach could reduce the SMO's time complexity from quadratic to linear in the number of input tables.

\subsection{Ablation Study}
\label{subsec:ablation-study}
In this section, we perform an ablation study of the usage of self-consistency to construct consistent workflow outputs and the usage of the LLM review loop in the Schema Matching Operator.

\begin{table}
    \centering
    \caption{Evaluation of the mappings output from the best-performing workflows per model when applying self-consistency (Self-Cons.) and average of 3 individual runs (Avg. 3 runs). Standard deviation is reported for the latter.}
    \label{tab:self-consistency}
    \resizebox{\columnwidth}{!}{
    \begin{tabular}{lrccc}
         \toprule
         \textbf{Model} & \textbf{Setup} & \multicolumn{3}{c}{\textbf{ Mappings to Integrated Schema}}\\
         & & \textbf{P} & \textbf{R} & \textbf{F1} \\
         \midrule
         \multirow{3}{*}{\textbf{Qwen-27B}} & Self-Cons. & 0.87 & 0.81 & 0.84 \\
         & Avg. 3 runs & \makecell[c]{0.74\\($\pm0.08$)} & \makecell[c]{0.68\\($\pm0.08$)} & \makecell[c]{0.71\\($\pm0.08$)}\\\midrule
         \multirow{3}{*}{\textbf{GPT-5.2}} & Self-Cons. & 0.86 & 0.83 & 0.84 \\
         & Avg. 3 runs & \makecell[c]{0.84\\($\pm0.04$)} & \makecell[c]{0.79\\($\pm0.03$)} & \makecell[c]{0.81\\($\pm0.03$)}\\
         \bottomrule
    \end{tabular}}
\end{table}

\subsubsection{Self-consistency.}
\label{subsubsec:self-consitency-res}
In Table~\ref{tab:self-consistency}, we report the evaluation of the mappings between input columns and integrated attributes that were output by the best-performing workflow per model on two setups: i) Self-Consistency: Each operator is run 3 times to build a consistent operator output before using this output as an input to the next operator. Only one \textit{consistent} output is generated per workflow in this setup. ii) Average of 3 runs: The workflows are run 3 separate times, which leads to 3 different outputs. We evaluate them and report their averaged results as well as the standard deviation of all metrics. As can be seen from the results listed in the table, when the workflow is run 3 separate times it produces results whose F1-scores vary with a standard deviation of $\pm0.08$ for Qwen-3.6-27B and $\pm0.03$ for GPT-5.2 (both models in this setup are run with a low temperature of 0.001 and 0, respectively, to reduce randomness). When we compare these results to the self-consistency setup, we can see that using the self-consistency strategy improves the F1-score by 13\% for the Qwen model, while in the more stable GPT-5.2 model the overall F1-score is improved by 3\%.

\begin{table}
    \centering
    \caption{Schema matching review loop effects on the mappings of the input columns to the intermediate integrated schema.}
    \label{tab:schema-matching-analysis}
    \resizebox{\columnwidth}{!}{
    \begin{tabular}{lrccccccc}
         \toprule
         \textbf{Model} & \textbf{Setup} & \textbf{P} & \textbf{R} & \textbf{F1} &\\
         \midrule
         \multirow{2}{*}{\textbf{Qwen-27B}} 
         & SMO + loop (Mappings) & 0.93 & 0.93 & 0.93 \\
         & SMO w/o loop (Mappings) & 0.82 & 0.82 & 0.82\\\midrule
         \multirow{2}{*}{\textbf{GPT-5.2}} 
         & SMO + loop (Mappings) & 0.86 & 0.84 & 0.85 \\
         & SMO w/o loop (Mappings) & 0.76 & 0.75 & 0.75 \\
         \bottomrule
    \end{tabular}}
\end{table}

\subsubsection{Schema Matching Operator Reviewer Loop.} In Table~\ref{tab:schema-matching-analysis}, we report the Precision, Recall and F1-score of the mappings of the input columns to the integrated attributes present in the intermediate integrated schema. These integrated attributes have been chosen from the created attribute groups ($AG$) formed using the Schema Matching Operator (SMO). In the table, we report the results when the attribute groups are refined using the SMO Reviewer Loop (SMO + loop) and when the Reviewer Loop is not used (SMO w/o loop). From the results, we notice that the Mapping F1-score increases by at least 10\% for both models. The Recall metric increases for both models when the loop is used, indicating that without the review, some false positives led to the wrong merging of attribute groups, which in turn led to missing integrated attributes and incorrect column mappings. In Table~\ref{tab:smo-loop-errors}, we show some correct and incorrect match removals.

\begin{table}
    \centering
    \caption{Correct/incorrect match removals for GPT-5.2.}
    \label{tab:smo-loop-errors}
    \resizebox{\columnwidth}{!}{
    \begin{tabular}{r|l}
\toprule
        & \textbf{Error pairs and some of their values } \\\midrule
        \multirow{1}{*}{Correct removal} & -- \textit{ParkUrl}: \{chicagoparkdistrict.com/parks/...\}\\ 
        & -- \textit{EventUrl}: \{chicagoparkdistrict.com/events/..\}\\
        \multirow{1}{*}{Correct removal} & -- \textit{VEI\_Holoce}: \{3, 5, ...\}\\
        & -- \textit{vei\_category}: \{Colossal, Super-Colossal, ...\} \\
        \multirow{1}{*}{Incorrect removal} & -- \textit{sTheaterAddress}: \{74 Warrenton Street;\\ 
        & 3790 Las Vegas Blvd S; ...\}\\ 
        & -- \textit{address}: \{1408 Locust, Des Moines, IA;\\
        & I -80, Exit 142, Altoona, IA 50021; ... \}\\ 
        & Reason: Due to partial information. \\
        \multirow{1}{*}{Incorrect removal} & -- \textit{closes}: \{Open Run; Open Run; ...\}\\ 
        & -- \textit{endDate}: \{2011-06-22; 2011-09-02...\}\\
        & Reason: Different value representation.\\
         \bottomrule
    \end{tabular}}
\end{table}

\section{Attribute Grouping Baselines}
\label{sec:baselines}
\textbf{Baselines.} To the best of our knowledge, no other works on schema integration or holistic schema matching divide columns of a single input table based on entity types. Therefore, we can not compare the full workflow, including table splitting, to baselines. As attribute grouping is a key operation in the SINT-Flow framework for generating the integrated schema, we compare the performance of the Schema Matching Operator with that of existing attribute clustering approaches. We choose the following baseline methods: \textbf{1) COMA}~\cite{do2002coma} a well-known schema matching symbolic label- and instance-based matcher, which we run by using the Python implementation by Koutras et al.~\cite{koutras2021valentine}\footnote{https://github.com/delftdata/valentine}. \textbf{2) ALITE}~\cite{khatiwada2022integrating} an integration method that embeds columns using different models and uses these embeddings to cluster the columns and create attribute groups. In their work they demonstrate that using the TURL~\cite{deng2022turl} model to create the column embeddings achieves the highest performance among other models tested. In their default embedding generation, ALITE does not consider the column headers as they test their system under the scenario of no table metadata available. In our paper, however, when we generate embeddings for the ALITE system, we use the column headers and in addition to TURL, we test embedding the columns using the Qwen-3-8B embedding model\footnote{\url{https://huggingface.co/Qwen/Qwen3-Embedding-8B}}. \textbf{3) SI-LLM}~\cite{wu2025schema} an LLM-based method for inferring a hierarchical schema from tabular repositories, which includes entity types, their unified attributes and annotated relationships between entity types. While in this work entity types are detected within a single input table, this detection is limited to named entity columns and to the main entity types detected in other input tables, and columns within the table are not assigned/related to the detected sub-entity types. Because of this, similarly to the other two baselines, we evaluate on the creation of the attribute groups within the tasks without dividing them into entity types. We use the prompts proposed in the paper for \textit{attribute name inference}, which renames the column headers based on the main entity type of the table, and for \textit{attribute names resolution}, where the renamed column headers are asked to be grouped under unified concepts that form the attribute groups. We use the SI-LLM prompts with GPT-5.2.

\textbf{Benchmarks.} We run the selected methods on SINT-Bench on two setups: in the first, we input to the methods the \textit{original} tables, while in the second, we input the \textit{split}, entity tables. We additionally evaluate on the Real Benchmark~\cite{khatiwada2022integrating}, a collection of 102 real-world tables divided into 11 tasks with topics such as transactions, school reports, events, etc.

\begin{table}
    \centering
    \caption{Attribute grouping results using the Schema Matching Operator (SMO) and the baselines.}
    \label{tab:baselines}
    \resizebox{\columnwidth}{!}{
    \begin{tabular}{lrccc}
         \toprule
         \textbf{Benchmark} & \textbf{Method} & \textbf{Precision} & \textbf{Recall} & \textbf{F1} \\
         \midrule
         \multirow{6}{*}{\textbf{Real}}
         & COMA & 0.74 & 0.64 & 0.68 \\
         & ALITE+TURL & 0.87 & 0.58 & 0.68 \\
         & ALITE+Qwen-8B & 0.76 & 0.75 & 0.74 \\
         & SI-LLM (GPT-5.2) & 0.92 & 0.66 & 0.76 \\
         & SMO + Qwen-27B & 0.91 & 0.95 & 0.92 \\
         & SMO + GPT-5.2 & 0.94 & 0.98 & 0.96 \\\midrule
         \multirow{6}{*}{\textbf{\makecell[l]{SINT\\(Original)}}}
         & COMA & 0.62 & 0.61 & 0.60 \\
         & ALITE+TURL & 0.78 & 0.58 & 0.63 \\
         & ALITE+Qwen-8B & 0.74 & 0.76 & 0.73 \\
         & SI-LLM (GPT-5.2) & 0.98 & 0.86 & 0.91 \\
         & SMO + Qwen-27B & 0.91 & 0.99 & 0.94 \\
         & SMO + GPT-5.2 & 0.94 & 0.98 & 0.96 \\
         \midrule
         \multirow{4}{*}{\textbf{\makecell[l]{SINT\\(Split)}}}
         & COMA & 0.62 & 0.64 & 0.62 \\
         & SI-LLM (GPT-5.2) & 0.99 & 0.87 & 0.93 \\
         & SMO + Qwen-27B & 0.95 & 0.98 & 0.96 \\
         & SMO + GPT-5.2 & 0.96 & 0.99 & 0.98 \\
         \bottomrule
    \end{tabular}}
\end{table}

\textbf{Results.} We report the results in Table~\ref{tab:baselines}. From the F1-score results, we notice that COMA and ALITE using TURL embeddings reach similar results on the creation of attribute groups. COMA achieves higher results when instance data overlaps between tables, e.g., when categorical attributes such as car type or brand are present in the use-case tables. TURL performs better in the use cases where some data was gathered from the WikiDBs corpus. This can be attributed to TURL being pre-trained with Wikidata information. On the other hand, TURL reaches lower F1-scores in the use cases where numeric or date columns are involved, such as low scores on the GOBY and Games tasks, which include a high amount of sales information. ALITE combined with Qwen embeddings has similar disadvantages related to numerical and date columns; however, overall it is able to identify the attribute groups with a higher F1-score.

Comparing the three prompt-based LLM methods, we see high results on both benchmarks for the Schema Matching Operator (SMO) for both Qwen-3.6-27B and GPT-5.2, while for SI-LLM we see high results on SINT-Bench but lower results on the Real Benchmark. The tasks with lower F1 for SI-LLM are the ones with a higher number of numerical columns and lower structural overlap. These lower results compared to SMO using the same model can indicate that finding column correspondences in a separate step is beneficial. This is similar to the conclusion that we reached when comparing Workflow 2 which does not use the SMO and Workflows 1 and 3 which do (see Section~\ref{subsec:workflow-results}). Some errors made by SMO using GPT-5.2 are listed in Table~\ref{tab:smo-errors}. 
GOBY and Flights were the tasks with the lowest F1-scores and some of the errors were made due to different value representations and on columns that had similar surface forms but pointed to different semantic concepts. When comparing the results of the methods on SINT-Bench with the \textit{original} tables to the ones on SINT-Bench with the \textit{split} entity tables, we notice that on the latter all methods' F1 results increase by around 2\%. This can be attributed to the separation of the attributes under different entities, which makes them easier to compare semantically, leading to better attribute groups. As ALITE is a column-based approach, the results are the same for both setups.

\begin{table}
    \centering
    \caption{Schema Matching Operator false positives and false negatives made by GPT-5.2 on SINT-Bench.}
    \label{tab:smo-errors}
    \resizebox{\columnwidth}{!}{
    \begin{tabular}{r|l}
\toprule
        \textbf{Type of error} & \textbf{Error pairs and some of their values } \\\midrule
        \multirow{1}{*}{FP (Flights)} & -- \textit{ArrAirport}: \{RNI, YWA, ...\}\\ 
        & -- \textit{IATA\_code}: \{NTY, CVN, ...\}\\
        & Reason: While the values match, in this use case\\
        & they refer to two different concepts: one includes\\
        & all airport codes and the other includes only the\\
        & arrival airports for certain flights. \\
        \multirow{1}{*}{FP (GOBY)} & -- \textit{fullDates}: \{May 22 at 4:30, May 21 at 4:30, ...\}\\
        & -- \textit{times}: \{Every Monday 6:30PM, 7:30 p.m., 10 a.m.\\
        & -4 p.m., Friday, April 29, 2011 at 8:00pm, ...\} \\
        & Reason: Overlap in values.\\
        
        \multirow{1}{*}{FN (GOBY)} & -- \textit{subCategory}: \{Variety and Specialty Show, Plays\\
        & and Musicals, ...\}\\ 
        & -- \textit{type}: \{Classes and Workshops, Visual Arts and \\
        & Museums, ...\}\\ 
        & Reason: No value overlap. \\
        
        \multirow{1}{*}{FN (GOBY)} & -- \textit{showTime}: \{7 pm, 9:00PM, ...\}\\ 
        & -- \textit{time}: \{8:30 p.m., Every Tuesday 7-8pm, 7:30 p.m.\\
        & and 9:45 p.m., ...\}\\
        & Reason: Different value representations.\\
         \bottomrule
    \end{tabular}}
\end{table}

\section{Related Work}
\label{sec:related-work}
This section compares the proposed framework with related work. We include papers covering \textit{schema integration} as well as the tasks of \textit{holistic schema matching} and \textit{ontology learning}. 

\begin{table*}
    \caption{Comparison of SINT-Flow to related methods. }
    \label{tab:related-work-comp}
    \centering
    \resizebox{\textwidth}{!}{
    \begin{tabular}{rrcccccc}
    \toprule
         \multirow{2}{*}{\textbf{Method}} & \multirow{2}{*}{\textbf{Category}} & \textbf{Entity} & \textbf{Attribute} & \textbf{Attribute} & \multirow{2}{*}{\textbf{Values}}&\multirow{2}{*}{\textbf{Table Splitting}}\\ 
         & & \textbf{Correspondences} & \textbf{Correspondences} & \textbf{Names} & & \\\midrule
         Navathe et al.~\cite{navathe1982methodology} & Symbolic, semi-automatic & yes & yes & yes & no & no \\ 
         Spaccapietra et al.~\cite{spaccapietra1992} & Symbolic, semi-automatic & yes & yes & yes & no& no\\ 
         Janga et al. ~\cite{janga2013tabular} & Symbolic, automatic & no & yes & yes & no & no \\
         Das Sarma et al.~\cite{das2008bootstrapping} & Probabilistic, automatic & no & yes & yes & no& no\\
         He et al.~\cite{he2003statistical} & Probabilistic, automatic &  no & yes & yes & no & no \\
         Su et al.~\cite{su2006holistic} & Statistical, automatic & no & yes & yes & no & no \\
         Pei et al.~\cite{pei2006novel} & TF-IDF-based clustering, automatic & no & yes & yes & no & no \\
         ALITE~\cite{khatiwada2022integrating} & Embedding-based clustering, automatic & no & yes & no & yes & no \\
         SI-LLM~\cite{wu2025schema}  & LLM-based, automatic & yes & yes & yes & yes & no \\\midrule
         SINT-Flow (Ours) & LLM-based, automatic & yes & yes & yes & yes & yes\\
         \bottomrule
        \end{tabular}}
\end{table*}

\textbf{Rule-based Symbolic Methods.} Earlier methods for schema integration are manual or semi-automatic approaches based on integration rules and human interaction~\cite{spaccapietra1992,navathe1982methodology,batini1984methodology,batini1986comparative}. Batini et al.~\cite{batini1986comparative} define a framework to compare some early schema integration methods and establish the characteristics that an integrated schema should have: completeness, correctness, minimality, and understandability. Navathe et al.~\cite{navathe1982methodology} propose a semi-automatic approach where schemata are integrated using the class and attribute correspondences found by the similarity between their naming and definitions. At certain points, a designer is asked to confirm/review the connections found by the method. Spaccapietra et al.~\cite{spaccapietra1992} introduce a rule-based method that uses inter-schema correspondence assertions to link similar elements (entity types/attributes/references). \textit{Workflow 1} in our work is inspired by this paper's methodology, where we find entity type correspondences with the Table Grouping Operator and attribute-level correspondences with the Schema Matching Operator. 
Janga et al. ~\cite{janga2013tabular} propose a framework for schema discovery and integration of web tables, where they unify the schemata by finding inter-schema relationships based on column name similarity and number of columns per schema.

\textbf{Probabilistic/Statistical Methods.} Das Sarma et al.~\cite{das2008bootstrapping} introduce the concept of \textit{probabilistic integrated schema} where multiple possible integrated schemata are generated and associated with probabilities indicating the likelihood that they correctly describe the input schemata. He et al.~\cite{he2003statistical} propose a statistical holistic schema matching method by presenting the idea of the existence of a hidden schema that probabilistically generates the observed schemata. Following work by Su et al.~\cite{su2006holistic} presents a holistic schema matching method for web query interfaces relying on synonym discovery and attribute co-occurrences to find attribute matches. He et al.~\cite{he2004discovering} view the task of schema matching as correlation mining and use both positive and negative correlations between attributes of different schemata to find complex matching beyond 1:1 matches.

\textbf{Clustering-based Methods.} In this category, attributes are clustered and an integrated attribute can be derived from each attribute cluster~\cite{alshaikhdeeb2015integrating,wu2004interactive}. Pei et al.~\cite{pei2006novel} use TF-IDF to represent the attribute schemata (composed of their names and descriptions) as vectors and use the incremental K-Means clustering algorithm. In ALITE~\cite{khatiwada2022integrating}, agglomerative hierarchical clustering is used on embedded columns where the number of clusters is selected by finding the number that maximizes the Silhouette Coefficient. The columns are embedded using an iterative method, where a sample of rows is first embedded per column and then compared (via the Euclidean distance) to another sample of rows from the same column. If the distance is not above a threshold, the embeddings are merged until the threshold is met.

\textbf{LLM-based Methods.} Ji et al.~\cite{ji2025table} focus on tabular integration; however, they tackle the tasks of entity matching and data fusion and assume that schemata have already been aligned. Wu et al.~\cite{wu2025taxonomy} develop both the embedding-based clustering approach EmTT and the LLM-based approach GeTT for inferring a hierarchical schema of entity types derived from input tables. This work is expanded in SI-LLM~\cite{wu2025schema}, where in addition to finding the entity types of each input table, the attributes with the same entity types are clustered and an integrated schema is created for each type. Furthermore,  named entity columns are detected within each input table and assigned an entity type to annotate the relationships between different entity types. In our work, we do not limit the detection to entity types grounded in named entity columns but allow the detection of abstract types like Flight, Eruptions etc. and we additionally split the tables by assigning attributes dependent on each type.

\textbf{Comparison of the Methods.} Table~\ref{tab:related-work-comp} shows a comparison of selected discussed methods. We categorize the methods by type and degree of automation, and distinguish whether they derive explicit correspondences between entity types and/or attributes, if attribute names and/or instances are used as signals to discover correspondences, and if the method does entity-wise decomposition of input tables. SINT-Flow covers the table splitting aspect in contrast to all the other methods.

\textbf{Other Related Work.} Arora et al.~\cite{arora2023language} propose using LLMs to infer the schema of semi-structured heterogeneous data. Wang et al.~\cite{wang2025schemaagent} use LLM agents to design a database schema from user queries/specifications. Bongertmann et al.~\cite{bongertmann2025large} tackle the problem of integrating heterogeneous sources with LLMs; however, they assume the target schema to be known.  
Barret et al.~\cite{barret2024computing} propose a method to create dataset abstractions by representing datasets as label-directed graphs. The method classifies the nodes into collections based on their semantic types and detects the main entities and the relationships between them. Closely related to schema integration and schema inference is the task of \textit{ontology learning}~\cite{asim2018survey}, where LLMs have been used for generating ontologies from unstructured data~\cite{babaei2023llms4ol}. Closer to our paper, there are works that focus on ontology learning from structured data~\cite{ma2022ontology}. Kayali et al.~\cite{kayali2024mind} experiment with building an ontology from input tables using an iterative column-based approach. OntoGenix~\cite{val2025ontogenix} builds an ontology from a single CSV dataset in multiple steps by first generating information about the dataset, column descriptions and their data types, and asking an LLM to enumerate the classes, subclasses and their properties. Laskowski et al.~\cite{laskowski2025burr} further extend this in their paper to evaluate the BURR benchmark by generating the descriptions and the ontology for multiple tables. The difference from our work stands in the inclusion of tables from different sources, which requires the unification of their schemata. Nayyeri et al.~\cite{nayyeri2025retrieval} use LLMs to generate ontologies for a relational database by using and extending a core ontology and using generated descriptions of columns and tables.

\section{Conclusion}
\label{sec:conclusion}
This paper introduced SINT-Flow, a framework for constructing schema integration workflows that derive integrated schemata for the entity types identified in a set of input tables. The framework consists of five LLM-based operators that can be arranged into different schema integration workflows. The paper also introduced SINT-Bench, a schema integration benchmark that covers integration tasks that require the identification of up to 3 different entity types per task. The evaluation of SINT-Flow on the SINT-Bench tasks demonstrated that its operator pipelines can derive integrated schemata without human intervention. We show that GPT-5.2 achieves the highest results when entity types are detected individually for each input table in contrast to the Qwen-3.6-27B model which favors the detection of entity types on an integrated view of all tables. Both models achieve a Completeness rate of at least 97\% for entity type detection and at least 81\% for attribute detection, resulting in an F1-score of 84\% for mapping input columns to integrated schema attributes. We show that the inclusion of the Schema Matching Operator in the workflows is beneficial, as the implicit schema matching workflow had the lowest F1-scores compared to the others tested. Finally, incorporating self-consistency to improve the consistency of individual LLM operators' outputs increased the F1-score for both models by 13 percentage points for Qwen and 3 points for GPT-5.2, compared to the average performance across three individual runs.
From a broader data engineering perspective, our experiments demonstrate that LLMs, when integrated into workflows that decompose complex tasks into manageable subtasks, can perform end-to-end schema integration without human intervention.

\section{Artifacts}
\label{sec:artifacts}
All the evaluation files, code and data, including SINT-Bench, to replicate the experiments in the paper are found in the SINT-Flow GitHub\footnote{\url{https://github.com/wbsg-uni-mannheim/SINT-Flow}} repository.


\balance
\bibliographystyle{ACM-Reference-Format}
\bibliography{si}


\end{document}